\documentclass[fleqn,10pt]{wlscirep}
\usepackage[utf8]{inputenc}
\usepackage[T1]{fontenc}
\usepackage{ltablex}
\usepackage{booktabs}
\usepackage{caption}
\usepackage{geometry}
\usepackage{comment}
\usepackage{soul}
\keepXColumns
\newcolumntype{Y}{>{\centering\arraybackslash}X}
\usepackage{float}
\usepackage{graphicx}
\usepackage{xurl} % 必须在 hyperref 之前
\usepackage[justification=justified]{caption}
\usepackage{afterpage}

\usepackage{pdflscape}
\usepackage{longtable}
\usepackage{tabularx}
\usepackage{booktabs}
\usepackage{makecell}
\usepackage{array}
\usepackage{ltablex}
\keepXColumns
\usepackage{multirow}

\title{Towards Generalist Intelligence in Dentistry: Vision Foundation Models for Oral and Maxillofacial Radiology}

\author[1]{Xinrui Huang}
\author[2]{Fan Xiao}
\author[2]{Dongming He}
\author[2]{Anqi Gao}
\author[2]{Dandan Li}
\author[3*]{Xiaofan Zhang}
\author[4,5*]{Shaoting Zhang}
\author[2,6,7,8,9,10,11*]{Xudong Wang}

\affil[1]{Shanghai Jiao Tong University, School of Information Science and Electronic Engineering, Shanghai, 200240, China}
\affil[2]{Shanghai Ninth People’s Hospital, Shanghai Jiao Tong University School of Medicine, Department of Oral Craniomaxillofacial, Shanghai, 200011, China}
\affil[3]{Shanghai Jiao Tong University, School of Computer Science, Shanghai, 200240, China}
\affil[4]{University of Electronic Science and Technology of China, School of Mechanical and Electrical Engineering, Chengdu, 611731, China}
\affil[5]{Shanghai Artificial Intelligence Laboratory, Shanghai, 200232, China}
\affil[6]{Shanghai Jiao Tong University, College of Stomatology, Shanghai 200125, China}
\affil[7]{National Center for Stomatology, Shanghai 200011, China}
\affil[8]{National Clinical Medical Research Center for Oral Diseases, Shanghai 200011, China}
\affil[9]{Shanghai Key Laboratory of Stomatology, Shanghai 200011, China}
\affil[10]{Shanghai Research Institute of Stomatology, Shanghai 200011, China}
\affil[11]{Chinese Academy of Medical Science, Research Unit of Oral and Maxillofacial Regenerative Medicine, Shanghai 200011, China}
\affil[*]{Corresponding Authors: xudongwang70@hotmail.com, xiaofan.zhang@sjtu.edu.cn, zhangshaoting@uestc.edu.cn}
\begin{abstract}
Oral and maxillofacial radiology plays a critical role in dental healthcare, while the interpretation of radiographic images is highly dependent on expert experience and limited by the global shortage of well-trained professionals. 
Although recent artificial intelligence (AI) approaches have demonstrated potential, existing dental AI systems are constrained by single-modality focus, task-specific design, and heavy reliance on high-cost labeled data, limiting their generalization across diverse clinical scenarios. 
To address these challenges, we propose DentVFM, the first family of vision foundation models (VFMs) tailored for dentistry, capable of generating task-agnostic visual representations for a wide spectrum of dental applications. 
DentVFM leverages self-supervised learning on DentVista, one of the largest curated dental imaging datasets with around $1.6$ million dental multi-modal radiographic images from multiple medical centers, and comprises 2D and 3D variants based on the Vision Transformer (ViT) architecture. 
To address the gap in dental generalist intelligence assessment and the limitations of benchmarks, we establish DentBench, a novel comprehensive benchmark covering eight dental subspecialties and encompassing more dental diseases and imaging modalities, with a wide geographical distribution.
DentVFM demonstrates impressive dental generalist intelligence that can robustly generalize to diverse dental downstream tasks, such as disease diagnosis, treatment analysis, biomarker identification, anatomical landmark detection and segmentation. 
Experimental results show that DentVFM significantly outperforms supervised, self-supervised, and weakly supervised baselines, exhibiting robust generalization, superior label efficiency, and high scalability.
Furthermore, DentVFM presents the cross-modality diagnostic potential, enabling more reliable diagnostics than experienced dentists in scenarios where conventional imaging modalities are inaccessible and resources are limited.
DentVFM introduces a new paradigm for dental AI, providing a label-efficient, adaptable, and scalable vision foundation model to advance intelligent dental healthcare and bridge a critical gap in global oral healthcare.
\end{abstract}
\begin{document}

\flushbottom
\maketitle
\thispagestyle{empty}

% \noindent Please note: Abbreviations should be introduced at the first mention in the main text – no abbreviations lists. Suggested structure of main text (not enforced) is provided below.

\section*{Introduction}

Dentistry focuses on the prevention, diagnosis and treatment of oral and maxillofacial diseases and disorders (e.g., caries, periodontitis, malocclusion, etc.). 
Dentistry encompasses various subspecialties, such as orthodontics, pediatric dentistry, periodontics, endodontics, prosthodontics, oral and maxillofacial surgery. 
Oral health exerts a substantial influence on the physical and psychosocial aspects of wellness\cite{barranca2022your}. 
World Health Organization (WHO) Global Health Status Report\cite{whooralhealth} shows that oral diseases affect approximately 3.5 billion individuals worldwide, predominantly in middle and low income regions. 
Radiographic imaging is at the forefront of dental healthcare due to its non-invasive nature. 
However, accurate interpretation of radiographic images requires significant investment from dental experts and demonstrates considerable variability between observers based on clinical experience. 
The growing demand for dental experts, coupled with an insufficient supply of well-trained professionals, has been further exacerbated by aging populations\cite{poudel2024oral}. 
Artificial intelligence (AI), particularly the emerging vision foundation model (VFM), is considered a potential solution to address these challenges in dental healthcare. 
Here, we make the first attempt to introduce the idea of VFM to dentistry. 

\begin{figure}[H]
\centering
\includegraphics[width=0.85\linewidth]{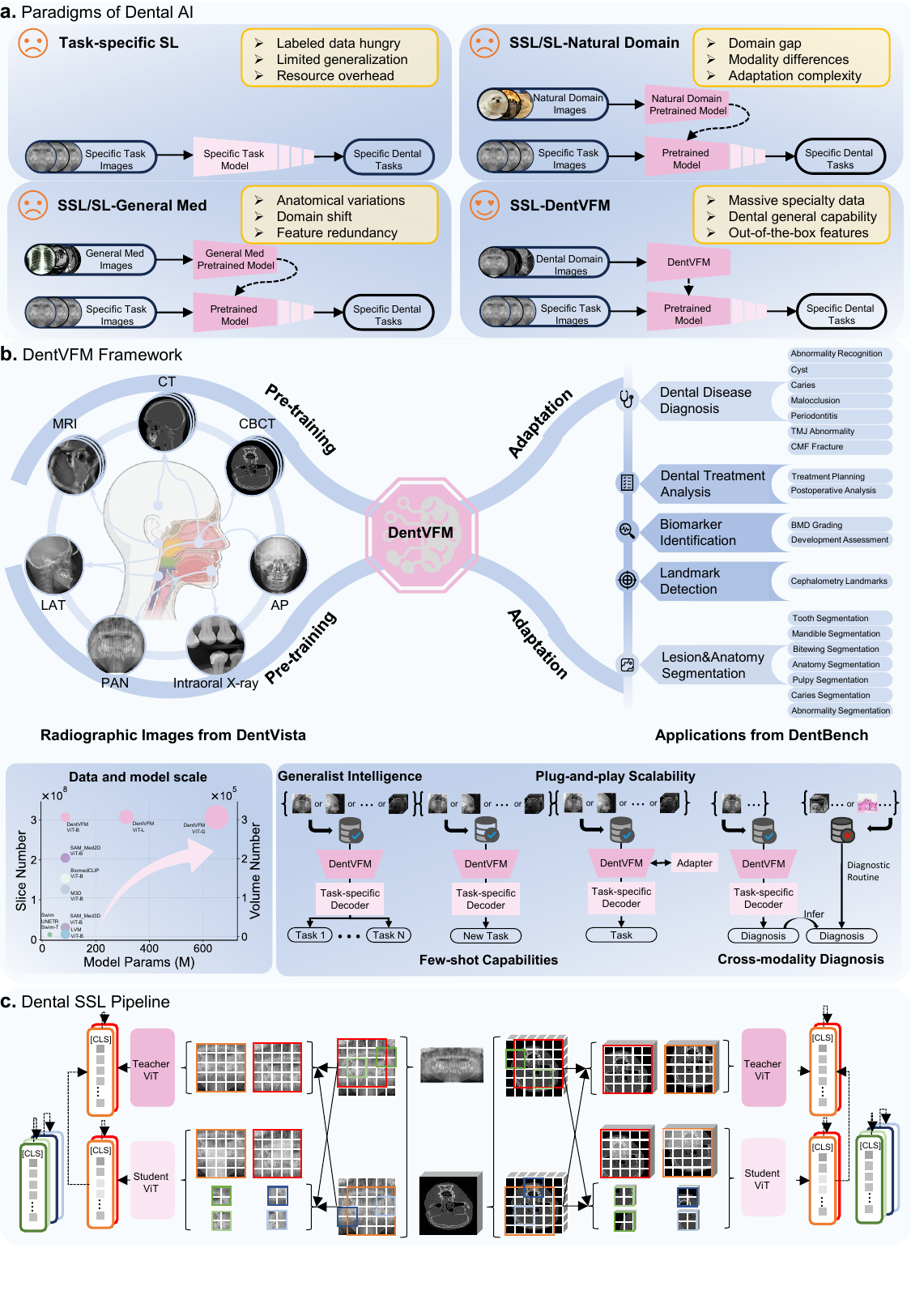}
\caption{Overview of the study. \textbf{a.} Different paradigms are employed in dental AI model development. DentVFM reconciles dental expertise with versatile applicability. \textbf{b.} DentVFM is built to be a multi-disease, multi-modal, multi-application foundation model using self-supervised learning based on the largest dental radiographic dataset, DentVista. It operates with larger model and data size. DentVFM exhibits generalist intelligence, few-shot capability, plug-and-play scalability, and surrogate modality diagnosis potential. \textbf{c.} The SSL method we employ is a self-distillation approach combining image- and patch-level objectives.}
\label{fig:main_framework}
\end{figure}

During the past decade, considerable effort has been made to develop conventional AI systems for dental radiographic image analysis \cite{li2022semantic,lang2022localization,hu2021location,wang2023multi,cui2022fully,mei2025clinical,lan2024mri,wang2023population,zhang2023deep,ito2022automated,huang2025maxillofacial}. 
Although progress has been made, some limitations still remain. 
First, dentistry involves a wide variety of multimodal radiographic images, along with their integrated analysis. 
Specifically, as the predominant examination technique, panoramic X-ray (PAN) provides comprehensive 2D visualization of oral and maxillofacial structures to diagnose dental diseases, e.g., impacted teeth, cysts, periodontitis, etc. 
Intraoral X-ray (e.g., periapical and bitewing imaging) is commonly used in endodontic and implant dentistry, offering detailed localized visualization. 
Anteroposterior and lateral X-rays (AP and LAT) are routinely employed in orthodontics and orthognathic surgery, serving as standard images for the assessment of maxillofacial deformity. 
Computed tomography (CT) and cone beam computed tomography (CBCT) can deliver 3D anatomical data, playing crucial roles in fracture diagnosis, implant planning, and orthognathic treatment. 
Magnetic resonance imaging (MRI), as a higher-cost imaging, is widely used in the diagnosis of soft tissue lesions, such as temporomandibular disorder (TMD) and tumors. 
Existing dental AI systems typically focus on analyzing a single modality and lack the ability to provide a unified feature extraction for processing multimodal images and combining multimodal information. 
Second, dentistry comprises multiple subspecialties and diseases and encompasses a wide range of application scenarios. 
Conventional dental AI models generally rely on specialized models designed for specific clinical tasks, focusing on a single or few dental diseases (see the Task-specific SL in Figure \ref{fig:main_framework}a). 
Developing specialized models creates significant operational overhead, and they have limited generalization to new diseases and new clinical applications. 
Third, these methods are data hungry for labels, requiring large volumes of high-quality labeled data, with annotations from experts being expensive and time-consuming. 
These limitations hinder the further application of AI in dental radiology. 

In recent years, self-supervised learning (SSL)\cite{he2022masked,oquab2023dinov2,bao2021beit,devlin2019bert,zhou2021ibot,caron2021emerging,balestriero2023cookbook,chen2021empirical} has been proposed to train models with transfer learning, generalization, and scaling capabilities through unlabeled large-scale data. 
In general computer vision, such self-supervised models have also been described as `vision foundation model' due to their generalist intelligence to adapt to a wide spectrum of downstream tasks\cite{bommasani2021opportunities,yuan2021florence}. 
In the medical field, various medical VFMs have also emerged as a focal point of research\cite{he2024foundation}. 
Depending on the constitution of their pre-training data, they can be simply categorized as modality-specific (e.g., X-ray\cite{tiu2022expert}, CT\cite{zhuang2025mim,huang2023stu,wang2023mis,zhuang2025advancing}, MRI\cite{luo2025large}, etc.) or organ/task-specific foundation models (e.g., computational pathology\cite{chen2024towards,hua2024pathoduet,wang2022transformer}, ophthalmology\cite{qiu2023visionfm,zhou2023foundation,cai2024uni4eye++}, endoscopy\cite{wang2023foundation}, etc.). 
Moreover, some studies improve the performance of VFM by integrating images and text, using weakly supervised signals derived from text\cite{blankemeier2024merlin,bai2024m3d,radford2021learning,wang2022medclip,zhang2023biomedclip}. 
VFM presents a promising opportunity to address the challenges faced by dental AI systems in a label-efficient, adaptable, and scalable solution that introduces a paradigm shift in the development of dental AI (refer to Figure \ref{fig:main_framework}a). 
However, generalizing existing VFMs to dental radiology is non-trivial. 
Adapting VFMs of the natural domain for dental radiology (as the SSL/SL Natural Domain in Figure \ref{fig:main_framework}a) presents a serious domain gap because radiographic images have unique characteristics whose modalities and patterns differ significantly from those of natural images\cite{zhang2024challenges}. 
Therefore, carefully designed adaptation algorithms are necessary. 
Existing medical VFMs are typically constructed based on organs with available large-scale public datasets, such as fundus, chest, abdomen, and brain imaging, leading to a large amount of redundant features and sparse dental knowledge. 
Transferring these models to dental radiology (as the SSL/SL General Med in Figure \ref{fig:main_framework}a) fails to achieve state-of-the-art performance due to the large variations present in organs and important structures, texture, shape, size, topology, and imaging modalities. 
Some attempts at constructing task-specific dental models have introduced self-supervised pre-training on dental data for initialization\cite{hu2021location}. 
However, these models suffer from limited pre-training data diversity, commonly restricted to small volumes of panoramic radiographs, leading to insufficient generalization across different dental radiological modalities and clinical tasks. 
As such, they fall short of the criteria for dental radiology foundation models with generalist intelligence. 

In this work, we aim to develop DentVFM, a novel family of vision foundation models for oral and maxillofacial radiology, which generates task-agnostic visual features that work out of the box on diverse dental applications, including disease diagnosis, treatment analysis, biomarker identification, anatomical landmark detection and lesion\&anatomy segmentation (see Figure \ref{fig:main_framework}b). 
DentVFM consists of DentVFM-2D, which focuses on 2D slices, and DentVFM-3D, which further considers the importance of perceiving the spatial semantics of volumetric images\cite{wang2024sam}. 
DentVFM applies the plain Vision Transformer (ViT)\cite{dosovitskiy2020image} as the foundational architecture and is constructed in multiple variants considering the scalability and deployment in hardware-constrained scenarios. 
DentVFM is pre-trained with SSL using one of the largest dental radiology collections, termed `DentVista'. 
DentVista is a pre-training dataset that consists of around 1.6M images (more than 30M slices), covering a wide spectrum of modalities, imaging devices, and demographics collected from 3 elite hospitals and 105 dental clinics (refer to Figure \ref{fig:dataset_statistic}a and b). 
Compared with conventional medical vision foundation models\cite{cheng2023sammed2d,zhang2023biomedclip,bai2024m3d,mh2023lvm,tang2022self,wang2024sam}, DentVFM achieves unprecedented advances in both data size and model size (refer to the Data and model scale in Figure \ref{fig:main_framework}b). 
The pre-training of DentVFM presents some distinct challenges, including managing dental multimodal data and selecting appropriate SSL algorithm. 
For data management, we establish a standardized pipeline for denoising, augmentation, and normalization of data from different modalities and protocols, producing suitable inputs for pre-training. 
For the selection of the algorithm, we apply the recently proposed DINOv2\cite{oquab2023dinov2}, which represents a more effective and memory-efficient discriminative SSL based on self-distillation. 
DINOv2 first augments input images to generate global and local crops, and then a pretext task is formulated combining both image- and patch-level objectives. 
DINOv2 can be seamlessly adapted to the pre-training of DentVFM-2D based on the standardized data pipeline, while more extensive customization is required for DentVFM-3D (more details refer to Figure \ref{fig:main_framework}c). 
Specifically, we replace the 2D tokenizer of ViT with a 3D tokenizer to accommodate volumetric data, and redesign augmentation strategies for dental 3D images. 

After SSL, DentVFM constitutes the first manifestation of generalist intelligence within dentistry. 
To assess the generalist intelligence, we constructed the `DentBench', a novel larger-scale dental radiology evaluation benchmark (refer to Figure \ref{fig:dataset_statistic}d). 
We strive to improve the comprehensiveness of DentBench by using extensively collected public dental datasets and carefully constructed complementary datasets. 
DentBench comprises five categories of downstream applications, more than 40 dental diseases derived from 8 subspecialties, covering 7 types of dental radiographic images sourced from 15 global regions. 
DentBench provides a wider spectrum of dental diseases and imaging modalities compared to previous dental radiology benchmarks\cite{wang2016benchmark,panetta2021tufts,hamamci2023dentex}. 
We apply different experimental settings on DentBench to validate the capabilities in multiple dimensions (see Figure \ref{fig:main_framework}b). 
Linear evaluation on DentBench demonstrates that DentVFM can learn robust universal representations capable of generalizing across heterogeneous dental applications and diverse dental diseases, thereby exhibiting characteristics of dental generalist intelligence (refer to Figure \ref{fig:overall_performance}). 
Compared to baseline models (based on supervised, self-supervised, or weakly supervised learning), DentVFM demonstrates superior performance across virtually all evaluated tasks, with particularly notable improvements observed in clinically critical applications including oral abnormality recognition, cyst diagnosis, temporomandibular joint (TMJ) abnormality diagnosis, and treatment analysis. 
Furthermore, DentVFM has exceptional few-shot learning capabilities, allowing generalization to new tasks and diseases with minimal annotated data (as in Figure \ref{fig:fewshot_performance}). 
Assessment in resource-limited settings proves that DentVFM attains a comparable performance to full labeled data training using only $25\%$ of the data. 
DentVFM also demonstrates high scalability and can serve as a plug-and-play module that seamlessly combines with parameter-efficient adaptation methods (e.g., linear adapter \cite{chen2021empirical,sowrirajan2021moco} for classification and ViTAdapter\cite{chen2022vision} for segmentation) and advanced task-specific frameworks (e.g., UNETR\cite{hatamizadeh2022unetr} and Mask2Former\cite{cheng2022masked}) as in Figure \ref{fig:head_to_head_performance}. 
Direct comparisons with existing task-specific models and experienced dentists reveal that DentVFM delivers comparable or even outstanding performance by constructing integrated models. 
In addition to exhibiting generalist intelligence, superior label efficiency and scalability, DentVFM presents a promising capability, cross-modality diagnosis, to mitigate the inequalities in dental healthcare resources. 
In dental clinical practice, clinicians routinely synthesize information from multiple imaging modalities for the diagnosis of complex diseases, for example, panoramic X-ray is not a conventional modality for the diagnosis of disorders of the TMJ, which require complementary MRI analysis. 
However, the high cost of certain imaging equipment (e.g., CT, MRI, pathological examination) prevents many resource-poor areas and small dental clinics from equipping all imaging capabilities. 
Consequently, patients are commonly limited to low-cost single-modal imaging such as panoramic radiography. 
In the TMJ abnormality diagnosis and cyst diagnosis tasks, DentVFM achieves substantial diagnostic precision using only panoramic X-rays without MRI and pathological examination, indicating that DentVFM has certain cross-modality diagnosis capabilities.

\section*{Results}

% \refstepcounter{section}
\phantomsection
\label{sec:result}
\addcontentsline{toc}{section}{Results}
\begin{comment}
此处总体描述结果部分：
在这个section中，我们设置了多种实验配方用于验证预训练dentvfm的有效性。
我们从提供一个关于dentvista和dentbench的总览开始。
Then，我们展示模型在dentbench上的评测结果。
具体来说，我们将dentvfm与10种其他预训练模型在fine-tuning设置下的性能进行比较，用于验证dentvfm的通用智能。
Following that，我们比较了dentvfm和其他方法在fewshot设置上的评测结果，用于验证模型在资源受限场景下的能力。
同时，我们在一些任务上执行了直接的头对头比较，将我们的方法与对应任务上报告的SOTA方法以及临床医生手动评测的结果进行比较，用于从临床经验角度评估模型。
另外，我们通过可视化表示和注意力图探索模型的可解释性。
最后，我们执行了消融实验以验证模型及预训练算法的合理性。
\end{comment}

In this section, we establish a comprehensive experimental framework to evaluate the efficacy of DentVFM. 
We begin with a statistical analysis of DentVista and DentBench. 
Subsequently, we present the evaluation results on DentBench, where we make comparisons between DentVFM and other pre-trained models using linear evaluation to directly assess the general capability of extracted feature. 
Following that, we investigate performance in few-shot settings to evaluate its label efficiency. 
Then, we conduct direct comparisons against task-specific models and clinicians to assess the scalability and cross-modality diagnostic capability from a clinical practice perspective. 
Additionally, we perform an ablation analysis to explore the scaling law in dental pre-training and the impact of pre-training configurations. 
Finally, we examine explainability through visualization of learned representations and attention mechanisms.

\subsection*{Statistics of Datasets}

We construct DentVista and DentBench datasets by collecting publicly available datasets, as well as data from top-tier dental hospitals and clinics.
DentVista is the largest oral and maxillofacial radiology dataset to date, designed for visual pre-training.
DentBench is a novel comprehensive evaluation benchmark that spans various dental diseases, subspecialties, and modalities to evaluate vision foundation models in dentistry.
To illuminate the characteristics of both datasets, we perform detailed statistical analyses as presented in Figure \ref{fig:dataset_statistic}.

\subsubsection*{Statistics of DentVista}

\begin{figure}[H]
\centering
\includegraphics[width=0.90\linewidth]{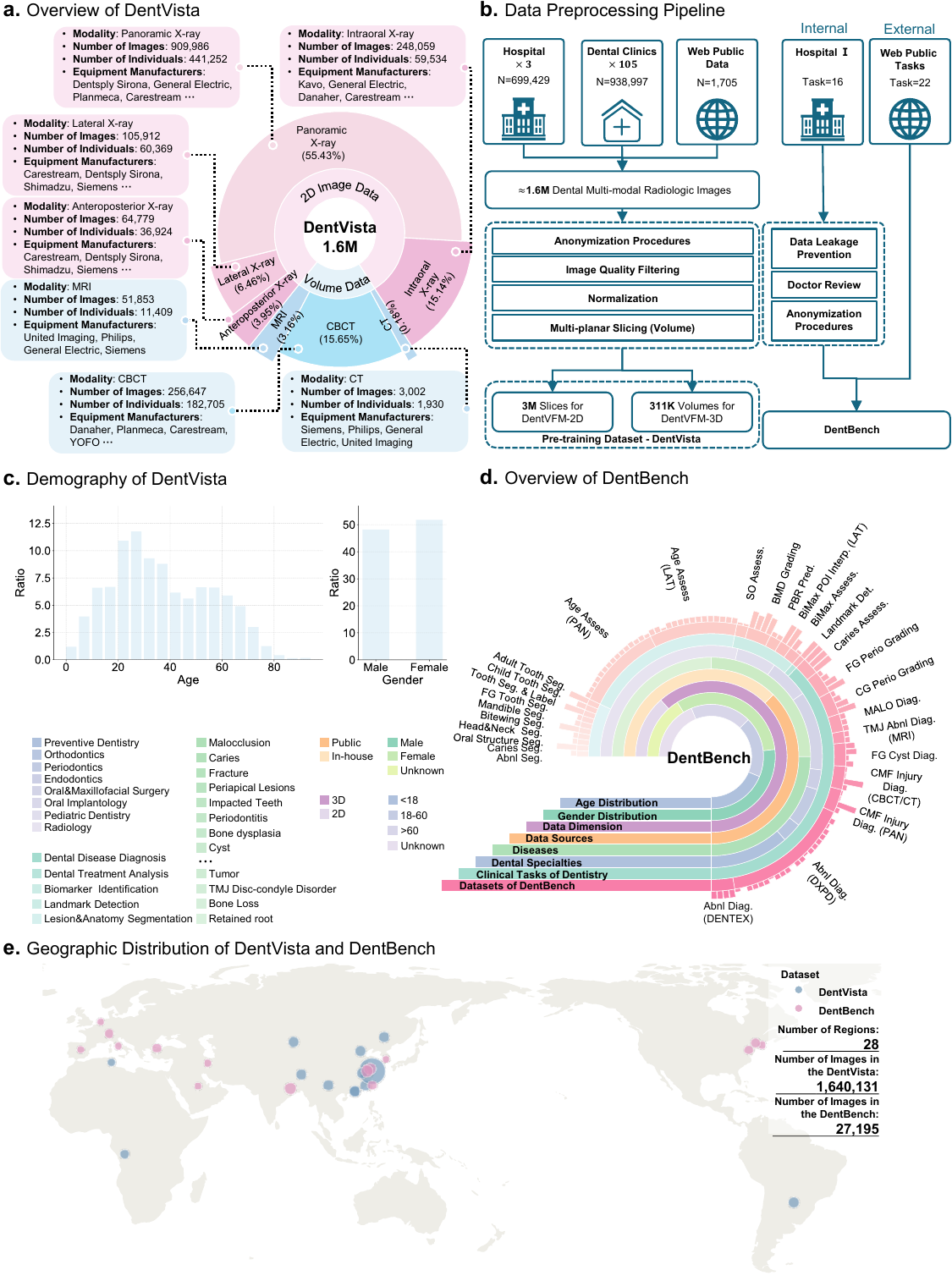}
\caption{Statistics of DentVista and DentBench. \textbf{a.} DentVista is the largest dental radiological dataset, covering 7 types of imagings obtained from various devices \textbf{b.} Data from multiple centers is preprocessed through customized pipeline to construct DentVista and DentBench. \textbf{c.} DentBase covers patients of all age groups and has an even gender distribution. \textbf{d.} DentBench consists of internal and external datasets, including 38 evaluation tasks across 5 clinical task types, 8 dental specialties, and more than 40 dental diseases. \textbf{e.} Our data covers diverse geographic locations (28 regions across 14 countries).}
\label{fig:dataset_statistic}
\end{figure}

Figure \ref{fig:dataset_statistic}a shows the compositional structure of DentVista, which comprises about 1.6M unlabeled multimodal images derived from 794K individuals in 13 regions (mainly in mainland China, detailed in Figure \ref{fig:dataset_statistic}d). 
DentVista data are collected from three dental hospitals and 105 dental clinics. 
The dataset incorporates 7 major types of multimodal oral and maxillofacial radiographic images (CT, CBCT, MRI, Intraoral X-ray, Panoramic X-ray, Anteroposterior X-ray, Lateral X-ray), captured by a wide range of devices. 
This improves the ability of the model to generalize across images from diverse imaging protocols, allowing it to effectively manage variations in spacing and field of view. 
Among these, panoramic X-rays account for the largest proportion (55.43\%), as panoramic imaging is a low-cost and commonly used dental radiographic examination technique. 
For volumetric data, CBCT represents the dominant share due to its widespread use in dentistry. 
The demographic analysis, shown in Figure \ref{fig:dataset_statistic}c, reveals that DentVista includes scans that span the entire age spectrum while maintaining a relatively balanced gender distribution. 
In contrast to most existing dental datasets composed of adult images, DentVista includes images from pediatric and geriatric populations. 
This broad demographic coverage allows DentVFM to be effective deployed in pediatric dentistry and elderly dental care. 

\subsubsection*{Statistics of DentBench}

The characteristic of DentBench is illustrated in Figure \ref{fig:dataset_statistic}d. 
DentBench is derived from 22 publicly available datasets and 16 carefully curated complementary datasets across 15 regions around the world. 
This benchmark contains oral and maxillofacial radiographic scans of more than 20,000 individuals, covering 8 dental subspecialties and more than 40 different pathologies. 
Through implementation of strict data isolation protocols between DentBench and DentVista, DentBench provides an ideal evaluation framework for interrogating the distributional robustness and generalization capabilities of pre-trained models across out-of-distribution (OOD) data. 
Specifically, public datasets can be considered as OOD tasks, since their data sources are not involved in the pre-training process. 
The downstream tasks within DentBench are systematically categorized into five fundamental categories—disease diagnosis, treatment analysis, biomarker identification, landmark detection and lesion\&anatomy structure segmentation—that span pivotal stages throughout the dental care continuum.
This systematic framework facilitates comprehensive evaluation of pre-trained models across the breadth of dental applications. 
More detailed descriptions of each task can be found in the Supplementary Table\ref{tab:dataset-summary}.

\subsection*{Performance of Model on DentBench}
To comprehensively assess DentVFM, we perform evaluations on DentBench. 
We used a full training set setting for general performance evaluation and a few-shot training set setting to assess the label efficiency of DentVFM.  
The evaluation results in both settings can be found in Figures \ref{fig:overall_performance} and \ref{fig:fewshot_performance}. 
More detailed evaluation results will be elaborated on later.

\subsubsection*{Evaluation of dental generalist intelligence}

We compare DentVFM with 11 baselines, which include models pre-trained on the general domain and medical domain, with different pre-training algorithms involving supervised learning, weakly supervised learning and self-supervised learning. 
Given different data dimensions, we evaluated the 2D and 3D versions of DentVFM separately. 

\begin{figure}[H]
\centering
\includegraphics[width=0.90\linewidth]{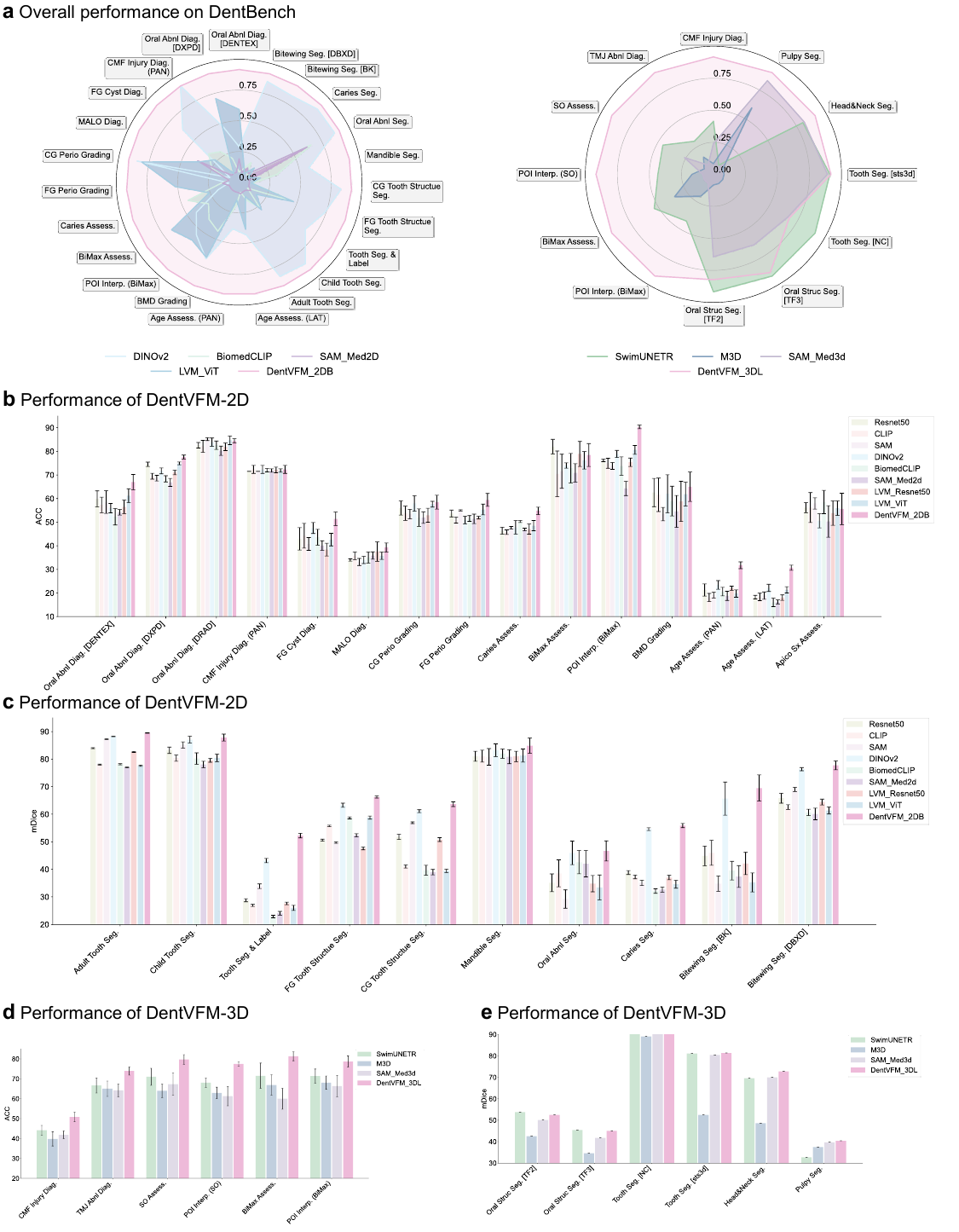}
\caption{Overall evaluation of the dental generalist intelligence. \textbf{a.} the overall performance of DentVFM on DentBench. 2D and 3D versions of DentVFM are assessed separately. \textbf{b} and \textbf{d} represent the performance of linear probing of DentVFM on 2D and 3D classification tasks. More pre-trained baselines are included, covering different architectures and pre-training algorithms. \textbf{c} and \textbf{e} are results of DentVFM integrated with lightweight segmentation heads on 2D and 3D segmentation tasks. DentVFM-2D is applied a linear segmentation head, while DentVFM-3D is integrated with a UNETR head. The error bars represent the standard deviation of 5 random train-test splits.}
\label{fig:overall_performance}
\end{figure}

The overall performance is shown in Figure \ref{fig:overall_performance}a. 
Here, we select baselines based on the same plain ViT architecture as DentVFM for fair comparison.
To evaluate native capabilities of pre-trained models, we employ lightweight classification and segmentation heads while keeping the pre-trained components frozen. 
Specifically, we apply a linear probe for classification (disease diagnosis, treatment analysis, biomarker identification), a linear segmentation head for 2D segmentation and a UNETR\cite{hatamizadeh2022unetr} head for 3D segmentation.
We perform five random data splits and report their average performance. 
This can be regarded as the overall performance on the dataset, thereby reducing the impact caused by random data splits. 
Figure \ref{fig:overall_performance}a shows that DentVFM outperforms all other baselines in diverse tasks. 
Furthermore, an analysis of baselines reveals that weakly supervised models (BiomedCLIP\cite{zhang2023biomedclip}, M3D\cite{bai2024m3d}) demonstrate better classification than segmentation performance. 
Here, we refer to downstream tasks such as disease diagnosis, treatment analysis, and biomarker identification, which can be formalized as tasks similar to classification, as classification.
In contrast, segmentation-specific models (SAM\_Med2D\cite{cheng2023sammed2d}, SAM\_Med3D\cite{wang2024sam}) trained with supervised learning exhibit better segmentation performance. 
Self-supervised models (DINOv2\cite{oquab2023dinov2}, LVM\_ViT\cite{mh2023lvm}, our DentVFM) support generalization to both classification and segmentation. 
Interestingly, DINOv2, the self-supervised model pre-trained on general domain datasets, has demonstrated superior performance across numerous dental tasks compared to LVM\_ViT, the previous model pre-trained on medical domain datasets.  
For more detailed explanations regarding all baselines, please refer to Supplementary Table\ref{tab:baseline_details}.
We also carry out a deeper investigation by comparing DentVFM with more baselines in Figure \ref{fig:overall_performance}b-e. 
Additional baselines include ResNet50\cite{he2016deep} (supervised pretraining on ImageNet), CLIP\cite{radford2021learning} (weakly supervised pretraining on WIT), LVM\_ResNet50\cite{mh2023lvm} (self-supervised pretraining on medical images) and SwinUNETR\cite{liu2021swin} (supervised pretraining on 3D medical images). 
DentVFM consistently outperforms other baselines in most tasks, highlighting its generalization in dentistry. 

Figures \ref{fig:overall_performance}b and \ref{fig:overall_performance}d provide a more detailed illustration of the evaluation results for classification tasks. 
The mean and standard deviation are shown for five random splits of the dataset
For the diagnosis of dental diseases, DentVFM achieves an improvement ($5.6\%$) in accuracy on \textit{OAR (DENTEX)}, a task focusing on recognizing four types of dental abnormalities (caries, deep caries, periapical lesions, and impacted teeth), compared to the second best method. 
It also delivers optimal results on \textit{OAR (DXPD)} and \textit{OAR (DRAD)} which cover more dental abnormalities. 
When it comes to the diagnosis of complex dental diseases, DentVFM also shows improved disease distinguishing ability. 
For example, \textit{CystDx} aims to diagnose confusable cyst types, including ameloblastoma, dentigerous cyst, keratocyst, and periapical cyst. 
DentVFM achieves an accuracy of 51.4\% using only a linear probe, significantly outperforming other pre-trained models (second best $47.5\%$). 
Similar performance gains are observed for other diagnostic tasks that require nuanced differentiation such as \textit{FG/CGPerioG} (periodontitis grading), \textit{TMJADx} (diagnosis of the TMJ disc displacement and changes in condylar position), \textit{CarA} (assessment of dental caries severity), \textit{CMFFxDx} (identification of craniomaxillofacial fracture sites) and \textit{MALODx} (diagnosis of malocclusion). 
For treatment analysis, we evaluate the capabilities of pre-trained models in orthognathic surgical planning as well as postoperative analysis. 
DentVFM shows better treatment analysis capabilities and has the potential to assist in planning and prognosis within clinical workflows. 
It achieves a precision of around $80\%$ in planning the required orthognathic surgical types based on preoperative scans (LAT or CT/CBCT) of patients with malocclusion (i.e. \textit{SOTP}, \textit{OSTP}), and an accuracy greater than 75\% in recognizing surgical types based on postoperative scans (\textit{SOPA}, \textit{OSPA}). 
For biomarker identification, we evaluated models on \textit{DevA (PAN/LAT)} (prediction of physiological age) and \textit{BMDG} (grading of bone density). 
DentVFM exhibits optimal performance on these tasks, demonstrating its remarkable proficiency in extracting subtle biomarker-associated characteristics from radiographic images. 

Figures \ref{fig:overall_performance}c and \ref{fig:overall_performance}e show the performance of the 2D and 3D versions of DentVFM in the dental lesion\&anatomical structure segmentation which plays an important role in dental treatment. 
For dental anatomical structure and restoration segmentation, we evaluated the capabilities of pre-trained models to segment critical oral structures such as teeth, jawbones, neural tube, or restorations from 2D or 3D scans. 
For lesion segmentation, we evaluated the performance in segmenting cavities or other abnormal lesions. 
As illustrated in the figures, DentVFM demonstrates superiority over other models in most tasks, underscoring the robustness and adaptability of learned representations, which can translate seamlessly from classification to dense prediction scenarios. 
The pre-trained model based on SwimTF performs well on some segmentation tasks due to the focus on visual inductive biases, but it reveals performance deficiencies in classification tasks.

\subsubsection*{Evaluation of label efficiency}
\begin{comment}
基础模型的一个重要优势是可以在少量标注数据下实现迁移学习。
为了系统的评价dentvfm的标签效率，我们在小样本的设定下评估模型泛化到新的下游任务的能力，即仅仅在训练中k%个标注样本。
因为few-shot学习对训练集的采样十分敏感，我们对每个k的取值进行了5次训练集的重新采样和模型重新训练并计算均值及误差带如图所示。
我们可以发现，训练数据越多，误差带越窄，性能也逐渐提高。
这种趋势在dentvfm2d和3d中表现一致，在分类任务和分割任务中表现一致。
令人惊讶的是，在一些任务上，如age、fgtooth、cGtooth任务上，dentvfm仅仅需要25%的有标注数据就可以达到100%标注相当的性能。
在与其他预训练模型的对比中，在大部分任务上，dentvfm仅需50%的标注数据即可达到甚至超过其他预训练模型在100%标注数据上的性能。
这些发现表明，dentvfm经过预训练获得了更强的牙科图像表征提取能力，在新任务的泛化过程中具有更好的标签效率。
【得益于大量牙科图像的预训练，dentvfm已经从中学习到从牙科图像中提取有表现力的表示的能力，在这种能力基础上的微调学习，仅需少量的标注数据即可让模型泛化到新的牙科任务上。】
\end{comment}

\begin{figure}[H]
\centering
\includegraphics[width=0.95\linewidth]{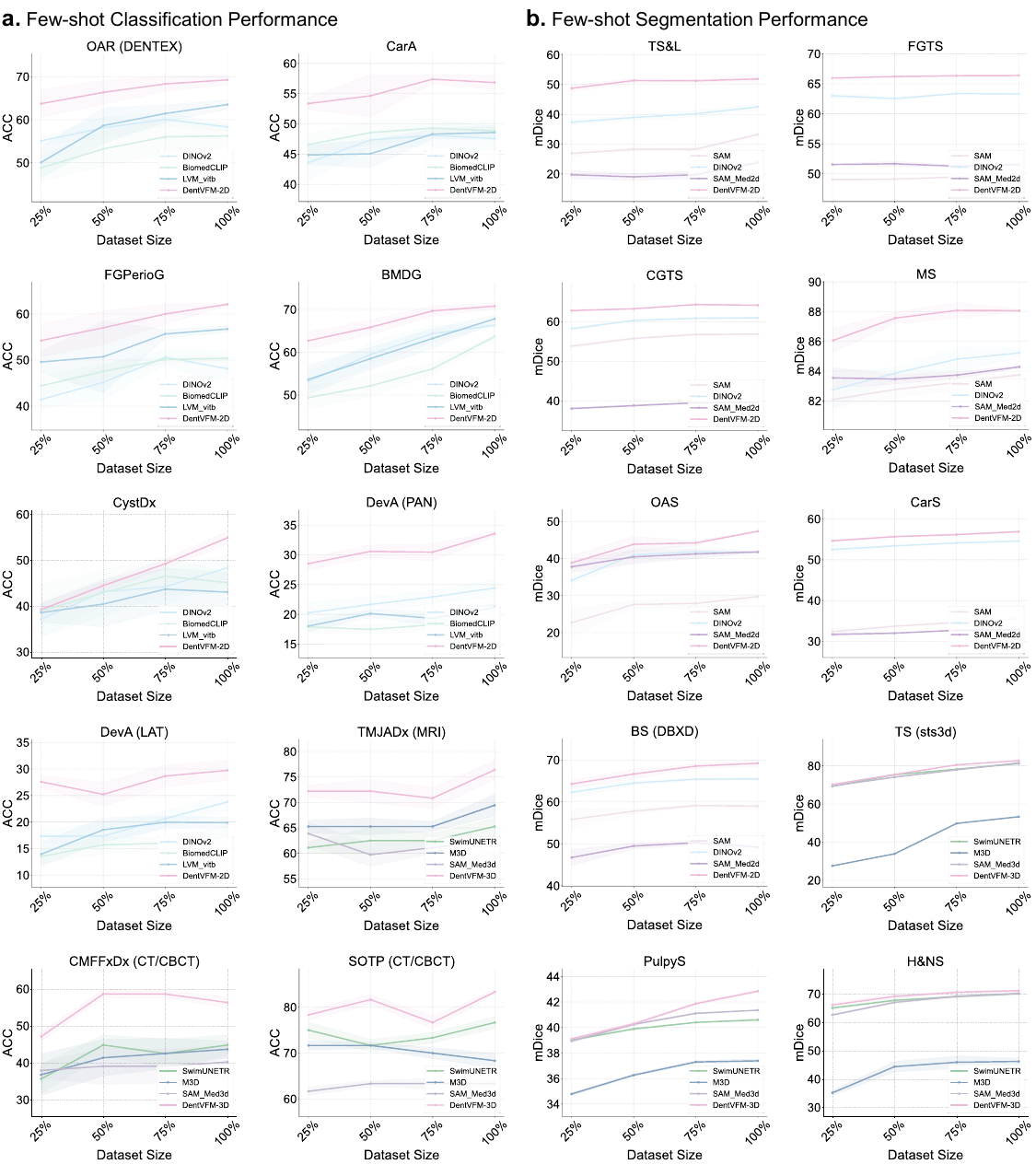}
\caption{Evaluation of the label efficiency of DentVFM. \textbf{a.} classification results with varying sizes of the labeled dataset, where the x-axis represents the training dataset size as a percentage of the total training dataset. The same test set is used for different percentages. Considering the impact of training set random samplings, we perform 5 samplings for each ratio and plot line charts with error bands based on the mean and standard deviation. \textbf{b.} segmentation results with multiple sizes of the labeled dataset. The classification and segmentation tasks we selected include tasks based on 2D and 3D images. We select competitive baselines from previous experiments for the corresponding tasks as a comparison.}
\label{fig:fewshot_performance}
\end{figure}

A key advantage of DentVFM is the ability to facilitate the adaptation of downstream tasks with few labeled data. 
To systematically evaluate the label efficiency of DentVFM, we perform evaluations in a few-shot setting, where only $k\%$ ($25\%$ to $100\%$) annotated samples are given during fine-tuning. 
Given the sensitivity of few-shot learning to the randomly sampled training data, we re-sample and re-train the model 5 times for each $k$ to calculate the mean performance and error bands, as illustrated in Figures \ref{fig:fewshot_performance}a and \ref{fig:fewshot_performance}b. 
We chose a set of pre-trained models that exhibit robust performance in prior evaluation as baselines here. 

As expected, the figures show that performance improves as more data is used for training, with narrower error bands. 
This trend demonstrates consistency across different DentVFM variants, as well as across classification and segmentation. 
Surprisingly, DentVFM, trained only with a small number of labeled examples ($25\%$), can achieve a performance comparable to training with the entire data set on several tasks such as \textit{DevA (LAT/LAT)}, \textit{OAR (DENTEX)}, \textit{CarA}, \textit{TMJADx (MRI)}, \textit{FG/CGTS}, \textit{MS} (mandible segmentation) and \textit{CarS} (caries segmentation). 
In comparative analyzes with other pre-trained models, DentVFM outperforms other models in terms of label efficiency, delivering comparable or even superior performance with $25\%$ annotated data required by competing models trained on full datasets. 
For example, DentVFM trained on $25\%$ of the labeled MRI achieves $72.22\%$ accuracy in TMJ abnormality diagnosis (\textit{TMJADx (MRI)}), exceeding the $69.44\%$ accuracy of the second best M3D fine-tuned throughout the data set. 
In the more challenging caries segmentation task \textit{CarS}, DentVFM demonstrates a Dice coefficient of $54.65\%$ using merely $25\%$ of the training data, while DINOv2 achieves only $54.61\%$ despite using the entire dataset. 
These findings suggest that DentVFM has learned diverse and expressive representations during pre-training, making it highly effective for new tasks even when finetuned on few labeled data.

\subsection*{Comparison with Specialist Models and Experienced Dentists}
\begin{comment}
为了对比dentvfm和专用模型的性能，我们选择了多个具有代表性的任务进行评测，涵盖分类任务（）如图a和分割任务（）如图b所示。
对于分类任务，我们在冻结的dentvfm后添加一个线性层，并在训练过程中执行数据增强，以增强模型性能。
对于分割任务，我们使用常用的adapter框架vitadapter，它被认为可以更好的将plain vit adapt到dense prediction任务上。
具体来说，我们在冻结的dentvfm上添加可训练的vit-adapter模块，并应用一个层次化的分割head。
我们对每个任务执行了K次随机划分并分别进行训练和评测以反映在数据集上的综合效果。

在囊肿诊断任务cyst上，dentvfm同专用模型lcd相比具有更高的acc、f1、auc。
在tmj诊断任务上，dentvfm同由imagenet初始化并完全fine-tune的resnet50相比表现更好。
为了从临床实践的角度对模型进行评估，我们还将dentvfm与具有不同临床经验的医生进行比较。
我们邀请了6名口腔放射科医生参与评测，其中3名医生为初级水平（从业1-3年），3名为中级水平（从业4-8年）。
医生根据共识进行手动评测，即只有两名及以上医生都认可当前放射图像属于同一种疾病，才会标记诊断当前疾病，否则将通过讨论直到得到共识。
如箱型图所示，在仅依赖放射图像进行诊断的设定下，我们的模型不仅优于专用模型而且优于两种类型的临床医生。
同时，混淆矩阵的分析说明我们的模型在易混淆的诊断类别上表现优于临床医生的手动评测。
对于龋齿分割任务caries，dentvfm相比unet和专用模型MLUA获得了更高的dice和iou。
对于从cbct中分割根尖病变区域任务（pal）以及分割72种口腔解剖结构的任务（tf3），dentvfm相比强的基线nnunnet也取得了提升。
对分割结果的可视化可以发现，在病灶分割任务中（caries、pal），dentvfm具有更好的目标区域完整性，在解剖结构分割任务种，dentvfm具有更好的目标语义准确性（比如牙齿的FDI编号识别）。
\end{comment}
DentVFM is highly scalable, which can serve as a plug-and-play module that integrates with parameter-efficient fine-tuning and advanced task-specific heads to enhance downstream performance. 
To assess its clinical practicality, we compare models integrated with DentVFM with specialized models. 
In addition, DentVFM exhibits cross-modal diagnostic potential, enabling accurate diagnosis using low-cost modalities for conditions that typically require complex imaging. 
This potential has significant implications for global oral healthcare.
It can facilitate disease screening in resource-constrained regions or facilities. 
In addition, it offers a promising approach to address the widespread issue of modality absence in the field of dentistry.
Compared with experienced dentists, we demonstrate the clinical potential of DentVFM.
All results are shown in Figure \ref{fig:head_to_head_performance}.

\subsubsection*{Evaluation of models integrated with DentVFM}
\begin{figure}[H]
\centering
\includegraphics[width=0.95\linewidth]{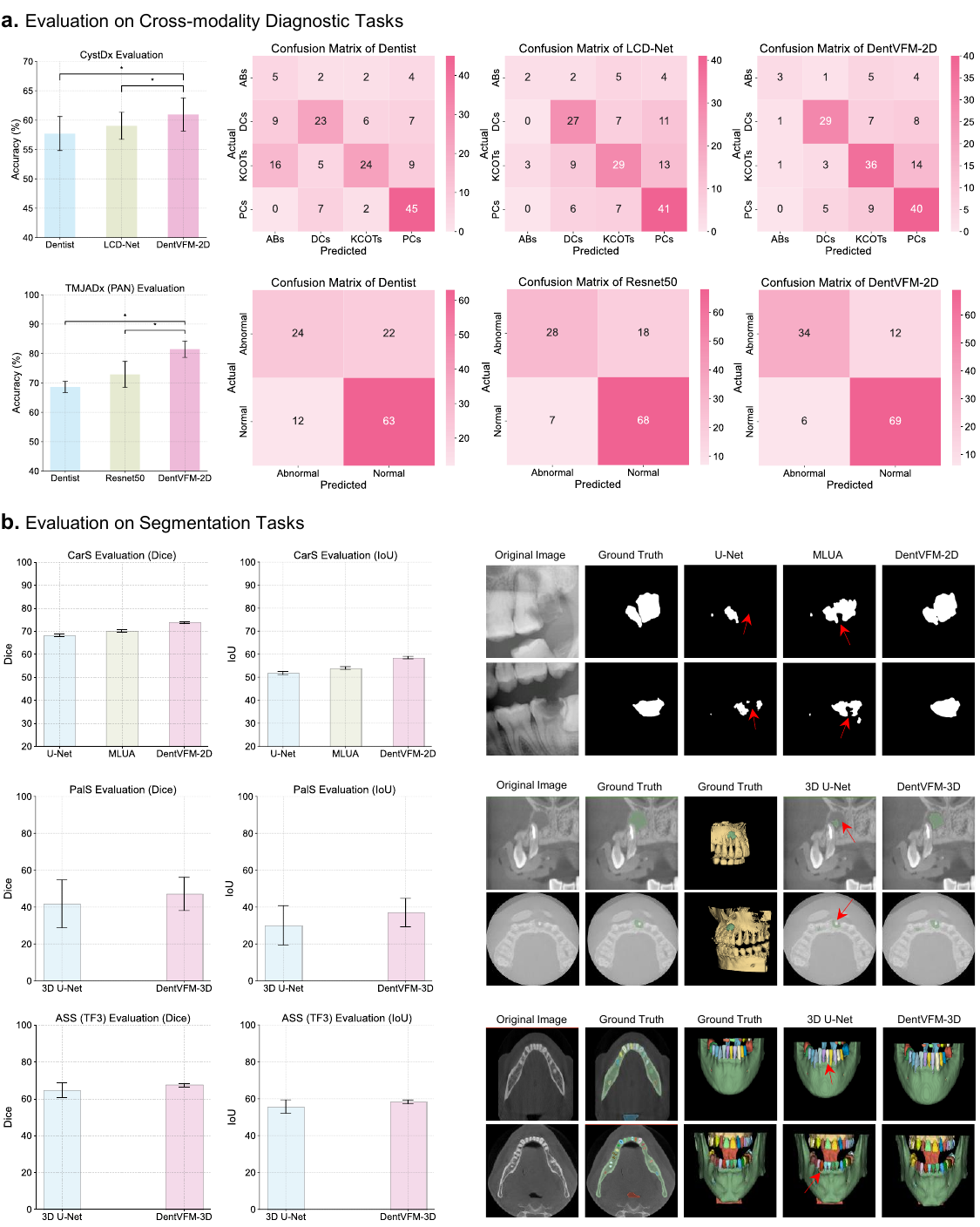}
\caption{Comparison with specialist models and experienced dentists. \textbf{a.} the accuracy and confusion matrices of DentVFM with linear adapters are evaluated on two cross-modal diagnostic tasks. We compare the performance with those of task-specific models and manual predictions made by experienced dentists. It demonstrates superior performance over specialist models in classification tasks when integrated with lightweight adapters, as well as more reliable cross-modal diagnostic capabilities than dentists. \textbf{b.} models integrated with parameter-efficient fine-tuning frameworks and DentVFM are evaluated on segmentation tasks. Quantitative and qualitative analyses show that constructing an integrated model can achieve better performance in segmentation tasks.}
\label{fig:head_to_head_performance}
\end{figure}

We perform extensive comparative analyzes between integrated models and specialized models in a set of five representative tasks, including classification tasks (Figure \ref{fig:head_to_head_performance}a), segmentation tasks (Figure \ref{fig:head_to_head_performance}b) and landmark detection task (Figure \ref{fig:ablation_study}d). 
For classification tasks, we add a trainable linear layer termed a linear adapter with frozen DentVFM and implement data augmentation. 
For segmentation tasks, we employ the widely adopted ViT-Adapter\cite{chen2022vision} framework, which has demonstrated efficacy in adapting a plain vision transformer for dense tasks. 
Specifically, we integrate trainable ViT-Adapter modules onto the frozen DentVFM and apply a hierarchical segmentation-specific head (UNETR\cite{hatamizadeh2022unetr} for DentVFM-3D and Mask2Former\cite{cheng2022masked} for DentVFM-2D) to generate predicted masks.
For the landmark detection task, we integrate frozen DentVFM with a trainable modified Mask2Former head, and use heatmap regression as the optimization target.

For the classification evaluation, DentVFM surpasses the specialized model, LCD-Net\cite{hu2021location}, in the diagnosis of cyst type (\textit{CystDx}) with statistical significance, as shown in Figure \ref{fig:head_to_head_performance}a. 
DentVFM also significantly outperforms fully fine-tuned ResNet-50 and achieves greater accuracy than $80\%$ for the TMJ abnormality diagnosis task (\textit{TMJADx (PAN)}). 
For segmentation evaluation, DentVFM achieves higher Dice coefficients and IoU scores compared to U-Net and MLUA\cite{wang2023multi}, the specialized segmentation model focused on caries segmentation, as shown in Figure \ref{fig:head_to_head_performance}b. 
In the periapical lesion segmentation task using CBCT data (\textit{ PalS }) and the segmentation task of the 77-class oral anatomical structure segmentation task (\textit{ASS (TF3)}), DentVFM shows improvements over the robust baseline U-Net under the same nnUNet\cite{isensee2021nnu} framework. 
We also visualize the predicted versus ground-truth masks for qualitative comparisons. 
Visual inspection reveals that DentVFM maintains superior lesion boundary integrity in pathological segmentation tasks, while demonstrating improved semantic fidelity in anatomical structure delineation, exemplified by precise FDI tooth classification. 
For anatomical landmark detection evaluation, DentVFM achieves better Mean Radial Error (MRE) and Success Detection Rate (SDR).
Visual analysis also indicates that the landmarks predicted by the model integrated with DentVFM have smaller deviations compared to the ground truth.

\subsubsection*{Evaluation of cross-modality diagnosis}
To evaluate cross-modal diagnostic performance, we select two complex diagnostic tasks, \textit{CystDx} and \textit{TMJADx (PAN)}, both characterized by the use of non-conventional imaging modalities for diagnosis. 
Specifically, \textit{CystDx} aims to perform a subtype classification of cysts using only panoramic radiographs from cyst patients, while in clinical routine, this differentiation typically requires further pathological examination. 
Similarly, \textit{TMJADx (PAN)} focuses on screening for disc displacement based solely on panoramic radiographs, in place of costly MRI examinations. 
DentVFM demonstrates considerable promise for cross-modal diagnostic inference. 
To benchmark diagnostic capabilities from a clinical practice perspective, we also performed comparative analyzes between DentVFM and experienced dentists with at least five years of clinical experience. 
Three oral oncology specialists are invited to perform manual evaluations on the CystDX task.
Three other dentists with experience with TMJ are invited to manually evaluate the TMJ task.
Dentists perform manual evaluations using a `consensus protocol', establishing diagnoses only after agreement by at least two dentists, and discordant cases resolved through discussion until consensus is reached. 
As demonstrated in the bar plots, DentVFM not only exceeds specialized models, but is also better than dentists, with an accuracy improvement of approximately $3.3\%$ (for \textit{CystDx}) and $13\%$ (for \textit{TMJADx (PAN)}), respectively. 
Furthermore, confusion matrices demonstrate that DentVFM outperforms manual assessment in diagnostically complex and ambiguous categories (e.g. DCs and KCOTs).

\subsection*{Ablation Analysis of Pre-training Configurations}
\begin{comment}
数据量、模型大小、预训练算法是构建具有通用智能的预训练模型的关键。
为了分析它们对牙科视觉基础模型预训练的影响，我们对模型进行了全面的消融实验，如图所示。

scaling law已经证明，在预训练中，增加数据集大小和模型参数量可以增强模型表现【】。
该现象不仅在自然语言中被观察到，在图像域也被观察到。
为了研究牙科放射图像领域中的scaling law，我们在不同大小的数据集上预训练了具有不同vit变体的dentvfm2d和dentvfm3d，并在多个下游任务上执行线性探查。
如图所示，相比使用少量数据预训练，纳入更多数据，可以显著提升预训练模型的能力。
同时，我们还观察到，在完整数据量的预训练下，放大模型规模可以实现持续的性能提升，但在数据量小的情况下，放大模型反而可能损害性能。
结果说明，要让更大的vit变体从预训练中获益，需要更多的预训练数据。
对牙科领域的scaling law研究凸显了利用更多数据取得卓越结果的潜力，进一步表明了多机构协作在聚合更大数据集以提高模型质量对于获得具有牙科通用智能的基础模型的价值。

为了分析不同预训练算法的影响，我们使用另一个被广泛用于医学影像预训练的算法（MAE）在dentvista上进行预训练。
对于2d影像我们使用vitb作为基础模型，对于3d影像我们使用vitl作为基础模型。
为了评价不同算法在dentbench上的整体表现，避免不同任务的数值差异，我们在每个任务上对不同算法训练的模型的预测指标进行归一化并计算所有任务的均值作为该预训练算法在dentbench上的平均表现。
如图所示，我们发现我们应用的dinov2算法显著优于MAE的预训练算法。
这证明了dinov2预训练算法在牙科放射领域的有效性。
【此处应该进一步扩充，什么东西可能是有效的，或者放到discussion中】

以往将预训练作为牙科放射学分析模型构建的一部分的工作一般都基于单一类型的牙科放射影像进行预训练比如全景片。
在我们的预训练过程中，我们将多种类型的影像混合在一起进行训练，比如dentvfm2d的训练过程中，我们的训练数据同时包含了全景、正侧位、牙片、CT slices、MRI slices。
为了研究混合数据对dentvfm的影像，我们从dentvista中过滤出所有全景片并基于单一影像对模型进行预训练。
我们选择了四个有代表性的任务，牙周炎分级、龋齿分级、术后影像术式识别、龋齿分割，进行评测。
如图所示，我们发现，使用混合数据预训练的模型的表现一致的高于只使用单一图像预训练的模型。
我们认为，这是因为不同类型的图像可以提供相同疾病的互补信息，并且这种互补信息会在预训练过程中被挖掘出来所导致的。
对于牙周炎的分级任务，该任务要求基于全景片，根据牙周病患者所展现出来的牙槽骨吸收程度对牙周病患者进行分级。
尽管分级诊断使用的输入是全景片，但是牙周炎患者的牙槽骨吸收模式可以在侧位片、正位片、牙片中被观察到，如图中黄色框所示。
这种互补信息有助于强化预训练过程中模型对牙周炎患者具有的异常牙槽骨的感知，从而在牙周炎分级任务中取得更好的效果。
对于龋洞分割任务，由于全景片中龋洞具有范围小，边界模糊的特点，仅在全景片上预训练的模型会面临挑战。
牙片主要关于局部几颗牙齿区域，具有更高的分辨率，更易于观察龋洞的形态、侵染范围，如图所示。
使用混合数据预训练的dentvfm由于可以从牙片中获得更多关于龋洞的互补信息，从而有利于在龋齿分割中取得更好的效果。
\end{comment}

Dataset size, model size, and algorithms constitute the fundamental pillars to build foundation models with generalist intelligence. 
Given the diverse range of imaging categories in dentistry, our DentVFM is specifically designed to undergo pre-training using a combination of imaging data.
We perform extensive ablation studies to assess our design and selection on dental pre-training, as shown in Figure \ref{fig:ablation_study}. 

\subsubsection*{Analysis of data and model size scaling}
Scaling laws have proven to be effective in improving the performance of foundation models by increasing the size of the training dataset and the model\cite{kaplan2020scaling}.
This phenomenon is observed not only in the natural language domain and in the image domain\cite{zhai2022scaling,dehghani2023scaling} but also in the medical domain\cite{zhou2023foundation}. 
To investigate scaling laws within the dental radiology domain, we pre-train DentVFM with different ViT variants and data size. 
We perform evaluations on multiple downstream tasks to demonstrate data and model scaling effects. 
As shown in Figure \ref{fig:ablation_study}a, the incorporation of more data during pre-training significantly improves performances. 
We also observe that scaling model size (from ViT-B to ViT-G) yields consistent performance improvements when pre-training with larger data size. 
However, model scaling may impair performance when pre-training with limited data (a subset of DentVista with 10k images). 
These results indicate that larger ViT variants require more data to benefit from pre-training effectively.
The investigation of scaling laws in the dental domain highlights the potential for achieving superior results by using more data, further emphasizing the value of multi-center collaboration in aggregating extensive data to construct a more powerful dental vision foundation model. 

\subsubsection*{Analysis of pre-training algorithm settings}
We analyze the impact of algorithm selection by employing another widely used medical pre-training algorithm\cite{zhou2023self,huang2024enhancing,cai2024uni4eye++,zhou2023foundation}, MAE\cite{he2022masked}, on the same DentVista as Figure \ref{fig:ablation_study}b. 
We utilize ViT-B as the base model for 2D images and 3D images. 
We report the normalized metrics of models pre-trained with different algorithms on each task and compute the mean across all tasks as the overall performance of DentBench. 
As demonstrated in the figure, DentVFM significantly outperforms MAE, validating the effectiveness of pre-training algorithm adopted by DentVFM.

\subsubsection*{Analysis of hybrid data utilization}
\begin{figure}[H]
\centering
\includegraphics[width=0.90\linewidth]{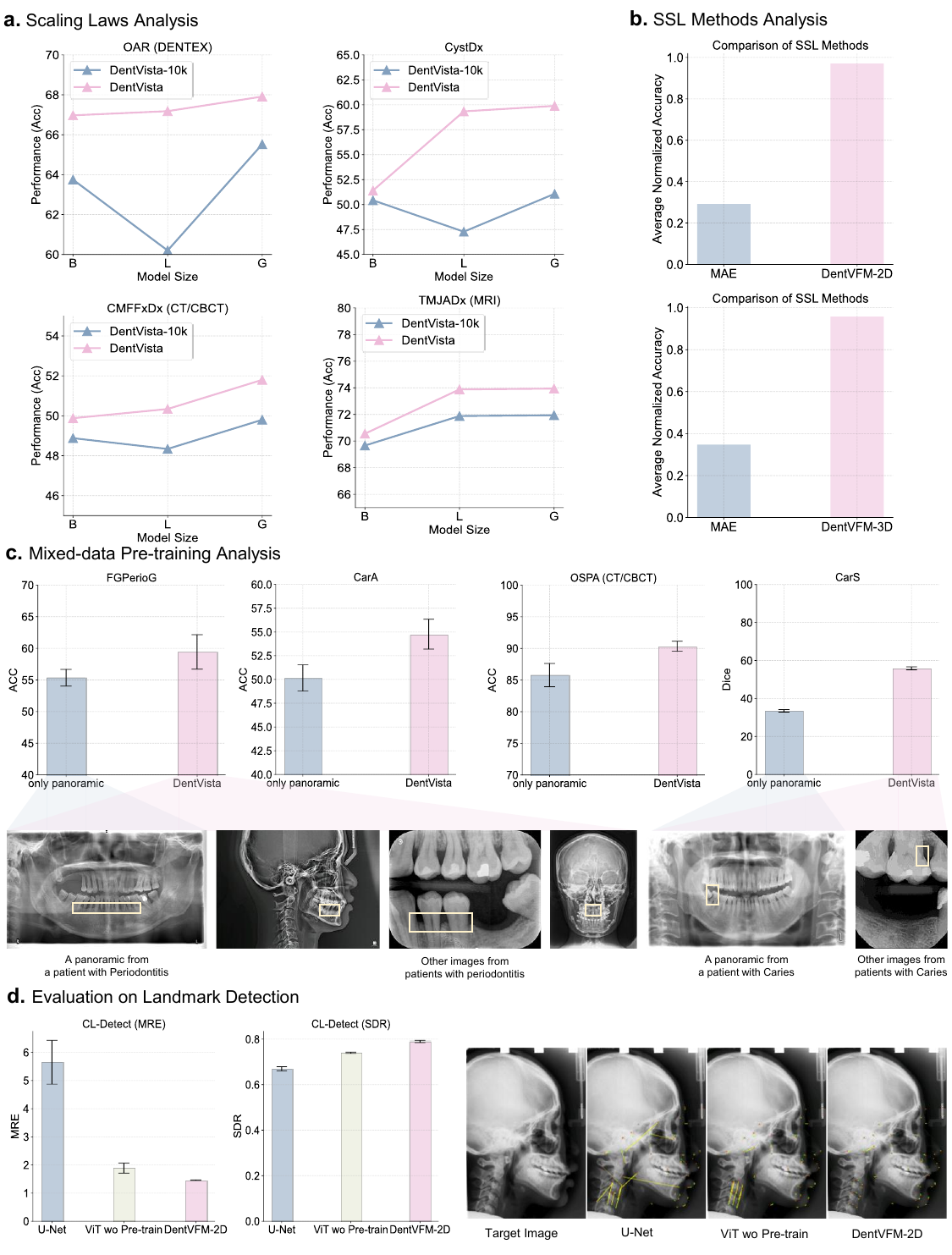}
\caption{Ablation analysis of pre-training configurations and an additional evaluation on anatomical landmark detection task. \textbf{a.} the scaling law of DentVFM, involving different model sizes (base, large and giant) and training data size. Both the data size and model size jointly impact the performance. Larger training datasets and bigger models offer more potential for dental image analysis. \textbf{b.} the impact of different pre-training algorithms on performance. \textbf{c.} applying hybrid multi-modal data to pre-training can leverage the complementary information. Four classification and segmentation tasks are selected to illustrate this. \textbf{d.} the evaluation of the model integrating DentVFM in locating key points in lateral X-ray images. The model integrated with frozen DentVFM and parameter-efficient fine-tuning methods achieves better performance at a lower cost compared to the fully fine-tuned U-Net and ViT based models.}
\label{fig:ablation_study}
\end{figure}

Previous dental pre-trained models typically conduct pretraining based on single types of dental radiographic images, most commonly panoramic X-rays. 
DentVFM is trained on hybrid imaging types, for example, DentVFM-2D is trained on data including panoramic X-rays, anteroposterior and lateral X-rays, intraoral X-rays, CT/CBCT slices, and MRI slices.
To investigate the impact of hybrid data, we filter all panoramic X-rays from DentVista and pre-train a new model based on these images. 
We select four representative tasks for evaluation as shown in Figure \ref{fig:ablation_study}c, i.e. \textit{FGPerioG}, \textit{CarA}, \textit{OSPA (LAT)} and \textit{CarS}.
As illustrated in the figures, the model trained in hybrid data consistently outperforms those trained in a single imaging type. 
We attribute this to the complementary information that different types of images provide for the same pathological conditions, which can be extracted during the pre-training process. 
The periodontitis grading task, \textit{FGPerioG}, requires grading periodontal patients based on alveolar bone resorption patterns observed in panoramic X-rays. 
Although the input to the task is panoramic images, alveolar bone resorption patterns can be observed in other types of images, that is, anteroposterior and lateral X-rays and periapical X-rays, as highlighted in the yellow boxes in Figure \ref{fig:ablation_study}c. 
This complementary information helps reinforce the perception of abnormal alveolar bone characteristics in periodontitis patients during pre-training, thus achieving superior performance.
For the caries segmentation task, models trained solely on panoramic radiographs face challenges due to the small extent and blurred boundaries of caries in panoramic images. 
Periapical radiographs focus on the localized tooth regions with higher resolution, which facilitates better observation of the morphology of caries and the extent of invasion. 
DentVFM pre-trained on hybrid data benefits from acquiring more complementary information about carious lesions from periapical radiographs, thus achieving superior performance in caries segmentation.

\subsection*{Explainability of Learned Representations}
\begin{comment}
dentvfm在各种牙科任务中展现出的通用智能来源于其强大的表示学习能力。为了更好的说明dentvfm是如何理解牙科放射图像的，我们对它提取的图像特征进行了不同粒度的可视化（包含图像级别、像素级别、体素级别）。为了分析dentvfm对图像整体的理解能力，我们选择了age和oral数据集进行图像级别表示的分布可视化。age数据集包含了处于颌面发育活跃阶段的未成年个体（6~18岁），并按相同年龄跨度（3岁）将其划分为四个年龄段。oral数据集包含了来自全景片的裁剪的ROI区域，带有四种口腔异常情况（龋齿、填充、阻生齿、种植牙）的标注。我们使用dentvfm2db对这两个数据集提取CLS token嵌入作为图像级别的表示，然后使用无监督的tsne对嵌入进行降维。如figurea中所示，来源于相同类别图像的图像级表示在空间中分布相近，形成一个个簇。在没有任何类别监督训练的情况下，dentvfm可以直接从图像中提取有意义的图像级判别性表示。【讨论，1、一些抽象语义概念，如年龄异常甚至性别等在自监督预训练过程中会被掌握，形成了一定程度开箱即用的能力；2、无监督训练的情况下提取有意义的判别性表示的能力对其在泛化到诊断类任务上的优良性能产生很大贡献】

为了进一步研究图像级判别性特征的来源，我们将dentvfm不同头部的注意力图可视化在不同模态的牙科放射图像上。如图figurec中所示，不同的注意力头部将关注图像的不同区域，来自所有头部的合并注意力图主要集中在对应模态图像中应被重点关注的区域。具体来说，对于正位和侧位xray图像，dentvfm关注前额骨、颧骨、上下颌骨、面部轮廓。对于全景xray和根尖xray，注意力主要集中于牙齿、牙槽骨及异常区域（如智齿、种植体）。对于CT和CBCT来说，dentvfm会关注牙列区域、脊柱、颌面部骨组织及软组织。对于mri，注意力集中于颞下颌关节盘及周围软组织。【讨论，通过可视化分析注意力图，我们发现dentvfm获得图像表示的方式具有与专业放射科医生读片过程相似的注意力模式，这展现了dentvfm具有合理性及一定程度的可解释性，这对于应用于医疗场景十分重要】
我们还在预训练过程中可视化了注意力图的演变。如图figured所示，随着学习的进行，dentvfm将逐渐关注到图像中更多关键区域，例如，对于侧位片，随着学习的进行，dentvfm将逐渐学会对上下颌骨的关注，对于全景片，dentvfm将逐渐提高对异常区域的关注，如上下的阻生智齿。【讨论，对学习过程的分析，展现了模型自主掌握这种注意力模式的动态过程，说明自监督学习可以帮助模型逐步掌握理解牙科图像的正确方式，模型可以逐步发现图像中应重点关注的区域】

为了探索dentvfm对牙科解剖结构的感知能力，我们对体素级别表示及像素级别表示进行了可视化。如figureb所示，我们选择了带有解剖结构分割标注的oral数据集。并使用dentvfm3db提取体素表示，并使用tsne方法可视化其表示分布。这些体素级别的表示形成了多个簇，并与不同的解剖区域相关联。我们还在figurec中使用kmeans聚类可视化了多模态数据的像素级别的表示。从图中可以看出，来源于同一解剖结构以及对称解剖结构的像素级别表示可以容易地被聚成一个cluster。【讨论，dentvfm可以自动学习到解剖结构对称性的概念无需额外训练，说明下人的对称性的生理学用处，以及关注对称性对某些任务的影响，以及对dense prediction任务的影响】
\end{comment}

The generalist intelligence demonstrated by DentVFM in diverse dental tasks stems from its powerful representation learning capabilities. 
To elucidate how DentVFM interprets dental radiographic images, we perform multi-granular visualizations of the extracted representations at different levels including image-level, pixel-level, and volume-level.

\subsubsection*{Visualization of image-level representations}
\begin{figure}[H]
\centering
\includegraphics[width=0.90\linewidth]{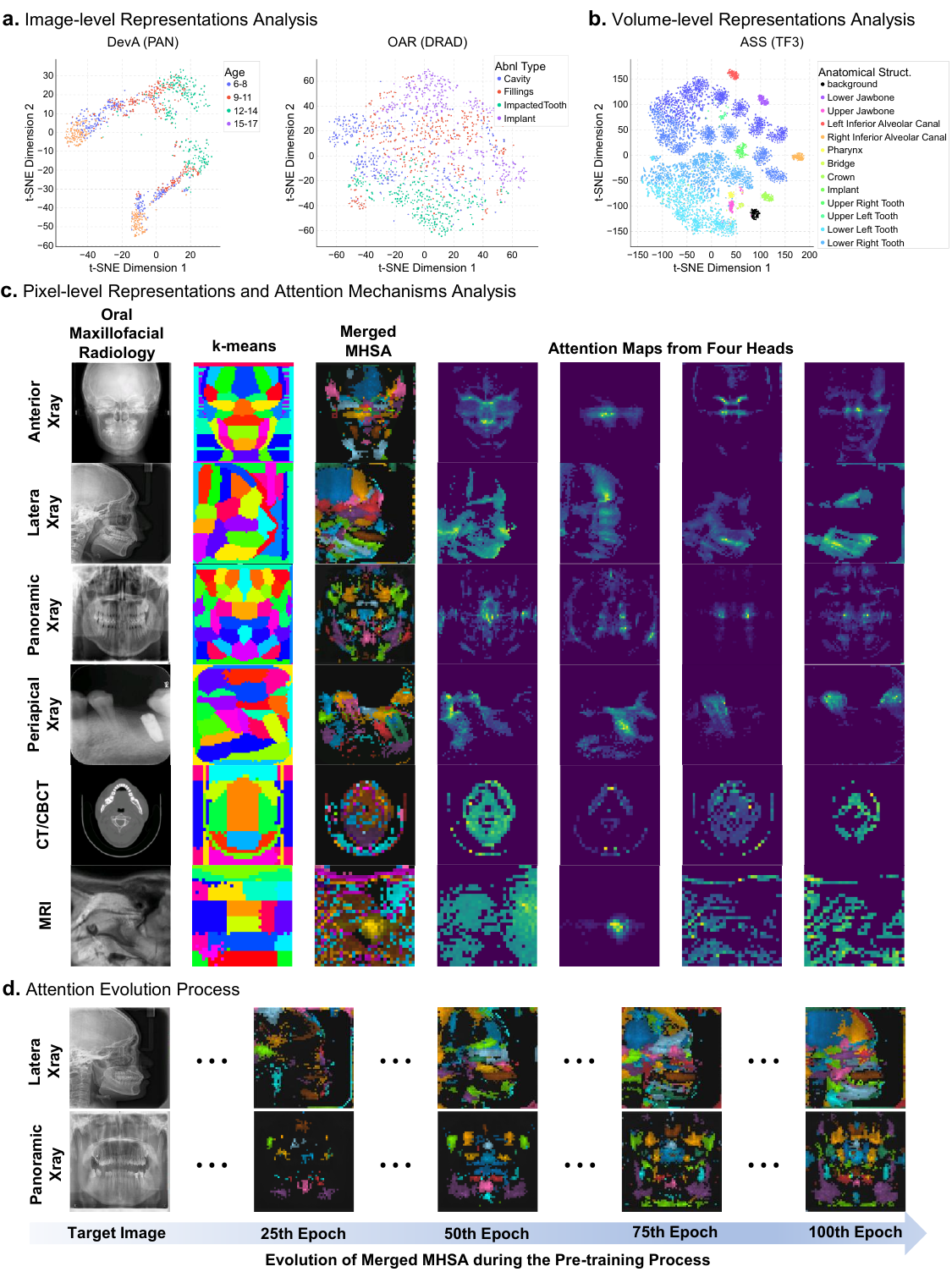}
\caption{Explainability of the learned representations of DentVFM. \textbf{a.} Visualization of the t-SNE projections of the learned image-level representations of DentVFM-2D on two typical downstream classification tasks. \textbf{b.} Visualization of the t-SNE projections of the learned volume-level representations of DentVFM-3D on a 3D segmentation task, demonstrating the anatomical awareness of the pre-trained model. \textbf{c.} Visualization of Multi-Head Self-Attention (MHSA) maps and pixel-level representations on images of different modalities. \textbf{d.} Visualization of the evolution of MHSA during pre-training, which demonstrates the performance enhancement brought about by pre-training.}
\label{fig:feature_visualization}
\end{figure}

To analyze the capacity of DentVFM for global image comprehension, we select the \textit{DevA (PAN)} and \textit{OAR (DRAD)} datasets for visualization of the distribution of image-level representations as shown in Figure \ref{fig:feature_visualization}a. 
The \textit{DevA (PAN)} dataset comprises adolescent subjects (aged 6 to 18 years) during an active phase of oral and maxillofacial development, stratified into four age groups with equivalent age intervals (3 years). 
The \textit{OAR (DRAD)} dataset contains cropped regions of interest (ROIs) from panoramic images, annotated with four types of oral abnormalities: caries, fillings, impacted teeth, and implants. 
We employ \textnormal{DentVFM-2D} to extract [CLS] token embeddings as image-level representations for both datasets and then apply unsupervised t-SNE \cite{maaten2008visualizing} to reduce the dimensionality of embeddings. 
Figure \ref{fig:feature_visualization}a shows that image-level representations corresponding to images within identical categories demonstrate spatial proximity in their distribution, resulting in the formation of distinctive clustering patterns.
DentVFM is capable of directly extracting semantically meaningful image-level discriminative representations from dental radiographic images without supervised training.

To further investigate the origins of image-level discriminative representations, we visualize the attention maps from different heads of DentVFM on images of different dental radiographic modalities. 
As illustrated in Figure \ref{fig:feature_visualization}c, different heads focus on different regions, and the merged attention map from all heads primarily concentrates on the region that should be emphasized in the corresponding modality. 
Specifically, for anteroposterior and lateral X-ray images, DentVFM attends to the frontal bone, zygomatic bone, maxilla, mandible, and facial contours. 
For panoramic and periapical X-ray images, the attention is predominantly focused on the teeth, alveolar bone, and pathological regions (e.g. wisdom teeth and dental implants). 
For CT and CBCT, DentVFM directs attention to the dentition, spine, maxillofacial bone structures, and soft tissues. 
For MRI, attention is centered on the temporomandibular joint disc and surrounding soft tissues.
We also visualized the evolution of attention maps during the pre-training process. 
As shown in Figure \ref{fig:feature_visualization}d, as pre-training continues, DentVFM will gradually attend to more critical regions of dental images. 
For example, DentVFM progressively learns to focus on the maxilla and mandible on lateral radiographs and increases its attention to pathological regions (e.g., impacted wisdom teeth) in panoramic radiographs.

\subsubsection*{Visualization of volume-level and pixel-level representations}
To probe anatomical region awareness in the embeddings of dental images, we perform visualization of volume-level and pixel-level representations. 
As shown in Figure \ref{fig:feature_visualization}b, we select the \textit{ASS (TF3)} dataset with anatomical structure segmentation annotations and employ \textnormal{DentVFM-3D} to extract volume-level representations. 
Then, we visualize their distributional patterns using t-SNE. These volume-level representations form multiple clusters that corresponded to distinct anatomical regions. 
We also visualize pixel-level representations of multimodal images using k-means in Figure \ref{fig:feature_visualization}c. 
As shown in the figure, pixel-level representations derived from identical anatomical structures, as well as symmetrical anatomical structures, are clustered into cohesive groups.
DentVFM can easily attribute clusters to anatomical regions and distinguish between different anatomical regions.

\section*{Discussion}
\begin{comment}
在本研究中，我们提出并验证了首个牙科视觉基础模型组 DentVFM，它通过在大规模多模态牙科放射影像数据集上的自监督预训练，展现出牙科通用智能。
为突破牙科公开数据资源有限的瓶颈，我们精心构建了 DentVista（迄今规模最大的牙科多模态放射影像预训练数据集）和 DentBench（一个覆盖更广、任务更全面的评测基准）。
DentVFM 包括专注于二维影像的 DentVFM-2D 与融合三维空间信息的 DentVFM-3D，两者在预训练后均表现出对多模态图像，多样化牙科任务优越的、高效的泛化能力。

在这个工作中，我们引入了第一个牙科基础模型组，dentvfm，它在大规模牙科多模态放射影像数据集上使用最先进的自监督学习算法进行训练。
为了突破牙科公开数据集数量少的限制以及对牙科任务进行更全面并且公平的评测，我们策划了dentvista（一个迄今为止最大的多模态牙科放射影像数据集）用于预训练和dentbench（一个更大、更综合的牙科放射评测基准）用于评测。
dentvfm由一个专注处理2D图像的dentvfm-2d和一个进一步考虑了3d空间信息的dentvfm-3d模型构成。
经过预训练后，dentvfm首次展现出了牙科通用智能。
它可以同时应对多种牙科放射图像类型并适配多种牙科下游任务涉及广泛的牙科疾病和应用类型。
将dentvfm适配到下游任务成本低，仅需少量标注数据和微调。
dentvfm具有高可扩展性，可以与现有的参数高效的微调框架进行集成，从而显著的改善ai模型在牙科任务上的性能。
我们还发现dentvfm展现出了令人惊讶的替代模态诊断的能力，这对缓解牙科保健资源不平衡，提高口腔疾病筛查效率方面具有重要意义。
作为一组被开发并验证的基础模型，dentvfm对在摆脱大量高质量标签限制的情况下构建牙科ai模型显示出相当大的前景。
\end{comment}
In this work, we introduce and validate the first family of dental visual foundation models, DentVFM, which demonstrates dental generalist intelligence through self-supervised learning on a large-scale multimodal dental radiographic dataset. 
To overcome the scarcity of public dental data and ensure a fair and comprehensive evaluation, we meticulously curate DentVista, the largest unlabeled multimodal dental radiographic pre-training dataset to date, and DentBench, a benchmark designed to evaluate broad and representative dental tasks. 
The DentVFM comprises DentVFM-2D, specialized in two-dimensional images, and DentVFM-3D, which incorporates three-dimensional spatial information, both of which achieved remarkable generalization after pre-training. 

\begin{comment}
dentvfm的牙科通用智能展现在多个维度。
dentvfm在跨越多种牙科放射影像类型、多种牙科应用类型、多种牙科疾病的下游任务上展现出显著的性能提升。
例如，DentVFM-2D在基于全景片的囊肿诊断、基于侧位片的畸形诊断以及基于咬翼片的结构分割任务中，分别较基线模型提升 11%、14% 和 7%；
DentVFM-2D在牙科疾病诊断应用，牙科治疗分析应用，牙科生物标志物识别应用，解剖结构和病灶分割应用中，分别较基线平均提升a%，b%，c%；
此外，DentVFM-3D在牙科治疗分析应用，牙科疾病诊断应用，解剖结构和病灶分割应用中，分别平均提升15%、7% 和 4%。
值得注意的是，性能提升同样出现在公开的牙科评测任务和补充构建的牙科评测任务中。
公开的评测任务可以被视为分布外的数据，因为提供它们的数据中心严格不会提供用于预训练的数据。
它们在地区、人种、成像协议的分布上同预训练数据有差异。
dentvfm在公开任务上相比基线平均提升a%，相对的，在补充构建的任务上平均提升b%，这表明其在分布外数据上的鲁棒性与普适性。
dentvfm相比在特定模态上预训练或者针对特定任务类型（比如分割）进行预训练的模型具有更强的通用性，更适合应对牙科诊疗面临的多模态影像处理、应用类型复杂的挑战。

dentvfm的牙科通用智能具有多维度。
具体来说，dentvfm提取的特征对多种牙科放射图像类型，多中心数据、多应用类型，多种finetune设置都具有很好的泛化性。
精心策划的包含了多维度的评测任务的dentbench验证了这种多维度泛化能力。
dentvfm可以用于不同类型的牙科放射影像特征提取，并提升后续分析的表现，比如，和baseline预训练模型相比，dentvfm-2d将基于全景片的囊肿诊断能力提升11%，将基于侧位片畸形诊断能力提升14%，将基于咬翼片的结构分割能力提升7%，dentvfm-3d将基于cbct的手术规划性能提升了15%，将tmj异常诊断能力提升了7%，将头颈解剖结构分割任务提升了4%。
我们用于评测模型的任务来源于多个医疗中心，包含了公开任务和内部整理的任务。
dentvfm在多中心任务上都取得了性能的改进。
值得注意的是，对公开任务的评测可以视为一种对应对分布外数据的能力的评估，因为用于预训练的数据严格地不包含这些任务中的任何数据。
dentvfm的通用性对ood数据也适用，比如，在牙科疾病筛查任务性能，牙齿分割标记任务上的性能提升，这些来源于公开评测任务的数据分布与预训练数据在地区、人种上的分布都有差异。
被用于验证dentvfm的通用能力的任务，从性质上可以简单的划分为两类，即分类和分割。
dentvfm的训练目标融合了image-level的目标和patch-level的目标，有助于同时提升预训练模型在两类任务上的效果。
图a中所示，不像现存的医学图像预训练模型要么关注分类任务要么关注分割任务，dentvfm的能力更平衡，在分类和分割任务上都有好的表现。
我们可以进一步观察到，dentvfm在分类任务上相比于基线模型的改进优于分割任务。
我们认为这主要是因为dentvfm的预训练应用的对比学习和自蒸馏方法强调multi crop中的特征一致性，缺少对像素级特征的一致性的建模，这强化了判别性高层抽象全局特征，其对分类问题友好，但对dense prediction任务的提升有限。
同时，相比于分类任务，分割任务微调过程中的训练数据集通常也包含了更多的样本，这一定程度上缩小了预训练的优势。
我们通过设定不同的微调样本数量比例，来模拟具有不同标注资源的场景。
dentvfm在不同few-shot设定下都显示出相比baseline更优的性能。
\end{comment}
DentVFM demonstrates remarkable dental generalist intelligence across multiple dimensions, showing substantial improvements across a range of downstream tasks involving multiple dental radiographic modalities, types of application, and diseases. 
For example, DentVFM-2D improves the second-best baseline models by $4\%$, $10\%$ and $4\%$ in cyst diagnosis based on panoramic X-rays, orthognathic surgery type identification using lateral X-rays, and structure segmentation utilizing bitewing X-rays, respectively. 
In tasks such as dental disease diagnosis, dental treatment analysis, biomarker identification, and anatomical structure and lesion segmentation, DentVFM-2D achieves average improvements of $3.5\%$, $6.8\%$, $6.7\%$, and $2.6\%$ over the second-best baseline models with the same model architecture. 
Furthermore, DentVFM-3D achieves average improvements of $13\%$, $8.5\%$ and $1.7\%$ compared with the second-best baseline models with the same model architecture in dental treatment analysis, dental disease diagnosis, and segmentation tasks.
In particular, performance gains are also observed in both public available evaluation tasks and additional custom-built evaluation tasks. 
Public tasks can be considered out-of-distribution (OOD) data, as the centers providing these task data do not offer any data used for pre-training. 
These tasks differ from the pre-training data in terms of regional, ethnic, and imaging protocol distributions. 
DentVFM outperforms baselines by an average of $2.5\%$ on public tasks, while showing an average improvement of $7.4\%$ on custom-built tasks. 
This highlights its robustness and versatility on OOD data. 
Compared to models pre-trained on specific modalities or task types (e.g., segmentation), DentVFM demonstrates superior generalizability, making it better suited to tackle the challenges of multimodal image processing and the complex application types encountered in dentistry. 

\begin{comment}
进一步的机制分析揭示，DentVFM的通用智能来源于提取的image-level强判别性特征的能力以及对牙科特定上下文的建模的能力。
预训练中使用的image-level目标和patch-level目标对这些能力产生贡献。
image-level目标强调不同增强view之间的cls token的分布一致性，这迫使dentvfm关注那些对标识相同图像起重要作用的判别性特征，比如和疾病、生物学标志相关的特征（如图a所示）。
进一步分析可以验证，image-level判别性特征来源于影像中的关键区域，比如特定解剖部位（比如，侧位片关注上下颌骨以及轮廓）和病灶区域（比如，口内xray关注牙齿及种植钉）如图b所示。
thanks to 这种对图像整体的良好建模，dentvfm可以在诊断、治疗分析、生物标志物识别等分类任务中取得显著改进。
patch-level目标通过掩码重建任务要求模型基于有限的可见图像块推断被遮蔽的图像块信息，使得dentvfm能够学习牙科影像的特异性上下文并对图像细节进行建模，从而提升在dense task上能力。
dentvfm的学习模式与临床医生的阅片逻辑存在某种程度的一致性，即分辨影像中的解剖结构，关注重点区域，识别异常情况类型，为模型在临床可解释性和可用性上提供了支持。

dentvfm的通用智能来源于对放射影像中强判别性特征的提取和对牙科特定上下文的建模。
dentvfm由image-level的目标以及patch-level的目标共同指导。
image-level目标令模型可以学习到判别性特征。
这种判别性特征既包括生物标志物方面的也包含疾病相关的。
TSNE分析，如图a，展示了dentvfm提取的原始无监督图像级别表示即可实现对生物标志（年龄）和口腔异常的状况的区分。
对图像级特征来源的分析可以发现其关注前景区域中的重点解剖部位，如侧位片中的上下颌骨、全景片中的牙齿区域、口内片中的异常牙齿区域等。
发育状态的推测和牙齿及颌骨密切相关。
口腔异常状况的诊断则需要强调对异常的区域的感知。
这是一种和临床医生阅片过程类似的影像解读模式。
patch-level目标使用MIM的代理任务让模型根据未遮盖部分推测被遮盖区域。
这种方式可以学习到牙科特定的上下文信息，有助于模型建模口腔影像中的解剖结构信息，
这为模型在分割任务中的性能做出贡献。
可解释性对于缓解将ai用于诊疗中的安全担忧具有重要作用。
dentvfm的可视化分析提供了一定程度的可解释性，这为其可以被集成于实际临床应用中提供了合理性。
\end{comment}
Mechanistic analysis indicates that the generalist intelligence of DentVFM derives from its capacity to extract effective discriminative image-level features and model the specific context of dentistry. 
Both the image-level and patch-level objectives employed during pre-training contribute significantly to these capabilities. 
The image-level objective prioritizes the consistency of the distribution of the [CLS] token in augmented views, compelling DentVFM to focus on critical discriminative features essential for identifying the same image, such as disease markers and biological indicators (see Figure \ref{fig:feature_visualization}a). 
Additional analysis reveals that these discriminative image-level features originate from key regions within the image, such as specific anatomical sites (e.g., lateral radiographs emphasizing the maxilla, mandible, and contours of the soft tissues) and lesions (e.g., intraoral X-rays that focus on teeth and implants), refer to Figure \ref{fig:feature_visualization}c.
Through comprehensive image modeling, DentVFM shows notable improvements in classification tasks, including diagnosis, treatment analysis, and biomarker identification. 
The patch-level objective, a mask image modeling pretext task, requires the model to infer information about masked image patches based on visible ones. 
This enables DentVFM to acquire a deep understanding of the dental context and model intricate image details, thereby enhancing its performance in dense tasks requiring fine-grained analysis, such as segmentation.
The DentVFM learning process is similar to, to some extent, the image interpretation strategies used by clinical professionals, who identify anatomical structures, focus on key regions, and identify abnormalities. 
This congruence improves the clinical interpretability and practical applicability. 

\begin{comment}
混合模态的预训练策略可以利用互补信息增强模型对牙科疾病的理解，进一步激发了模型的替代模态诊断能力。
通过对数据构成的消融分析，我们验证了混合数据预训练策略的有效性，使用混合数据训练的模型相比使用单模态预训练的模型在多个任务上都表现出一致的优越性能，如图b所示。
我们观察到，来源于患有同种疾病的患者的不同类型的影像数据可以提供不同角度的疾病模式形成互补信息。
例如，周炎患者所具有的牙槽骨质吸收的特征可以在侧位片、口内射线片，正位片中被观察到。
具体来说，全景片提供了关于水平骨质吸收信息，侧位片可以提供垂直向的骨质吸收信息以及牙齿松动度的评估，口内xray可以精细的反映局部牙-骨结构，牙周炎的分级也依赖于对局部的结构变化统计，正位xray可以提供牙周炎对全口影响的信息以及对颌骨的影响。
仅仅依赖全景片训练的模型无法融合来源于多类数据中共现特征的信息。
龋齿主要涉及牙齿的局部变化，尤其是牙釉质和牙本质的损坏。
由于分辨率的限制，初期龋齿及邻面龋齿通常非常难以从全景片中分辨出来。
口内X光片由于其高分辨率的特点，可以捕捉到更小的细节，并为模型提供关于龋齿的细粒度互补信息。
同样，来自3D 模态的slice数据的引入带来了类似的信息传递，可以激发模型使用替代模态数据进行诊断的能力。
例如，仅依赖全景片即可辅助筛查原本需要 MRI 的颞下颌关节盘异常。
这一能力为缓解资源匮乏地区的牙科影像学诊断困境提供了新的可能。

使用混合不同类型数据预训练的模型，可以利用互补信息增强对牙科疾病的理解，并进一步激发模型的替代模态诊断能力。
通过对数据构成的消融分析，如图b所示，我们发现使用混合数据训练的模型在多个任务上都表现出一致的优越性能。
我们认为，不同类型数据中关于同一个疾病的互补信息在其中扮演了重要角色。
通过对训练数据的进一步视觉分析我们发现，同样的疾病模式可以在不同类别的影像数据中被观察到。
如图b所示，牙周炎患者所具有的牙槽骨质吸收的特征可以在侧位片、口内射线片，正位片中被观察到。
全景片提供了关于水平骨质吸收信息，侧位片可以提供垂直向的骨质吸收信息以及牙齿松动度的评估，口内xray可以精细的反映局部牙-骨结构，牙周炎的分级也依赖于对局部的结构变化统计，正位xray可以提供牙周炎对全口影响的信息以及对颌骨的影响。
仅仅依赖全景片训练的模型无法融合来源于多类数据中共现特征的信息。
同样的，龋齿主要涉及牙齿的局部变化，尤其是牙釉质和牙本质的损坏。
由于分辨率的限制，初期龋齿及邻面龋齿通常非常难以从全景片中分辨出来。
口内X光片由于其高分辨率的特点，可以捕捉到更小的细节，并为模型提供关于龋齿的细粒度互补信息。
在训练数据中包含来自3d模态的slice数据也带来了类似的信息传递，可以激发模型使用替代模态数据进行诊断的能力。
通常被认为需要mri数据进行辅助诊断的颞下颌髁突关节盘关系异常，可以仅依赖作为替代模态的全景xray进行筛查，如图b。
这对缓解牙科保健资源的地域、机构分布不平衡带来曙光，毕竟昂贵的ct、mri设备是低收入地区及小型医疗机构所难以负担的。
\end{comment}
The pre-training with hybrid data capitalizes on complementary information to improve understanding of dental diseases, further stimulating the ability to diagnose based on surrogate modality. 
The ablation study of the data composition validates the effectiveness of this strategy as shown in Figure \ref{fig:ablation_study}c. 
Models trained on hybrid data consistently outperform those trained with single-modal pre-training across multiple tasks. 
Our analysis reveals that different types of images from patients with the same disease provide complementary insights by offering diverse perspectives on disease patterns. 
For example, alveolar bone resorption in patients with periodontitis can be observed on lateral X-rays, intraoral X-rays, and anteroposterior X-rays. 
Specifically, panoramic radiographs reveal horizontal bone resorption, lateral radiographs visually compare the height of the anterior and posterior alveolar bone, intraoral X-rays capture detailed local tooth bone structures, and anteroposterior X-rays can assess the symmetry of the left and right alveolar bone.
A model trained solely on panoramic radiographs will struggle to effectively integrate co-occurring features from multimodal data. 
Cavities mainly involve localized changes in the teeth, especially damage to enamel and dentin. 
Due to resolution limitations, early stage caries and interproximal cavities are often difficult to detect on panoramic radiographs. 
In contrast, higher-resolution intraoral X-rays provide more precise details and offer complementary fine-grained information on caries. 
Similarly, incorporating slices from 3D modalities enables further information transfer, enhancing the use of data of the surrogate modality to diagnosis. 
For example, panoramic radiographs alone can assist in the screening for abnormalities in the disk of the temporomandibular joint, which would typically require an MRI. 
This capability presents new opportunities to address diagnostic challenges in dental imaging, particularly in resource-limited settings. 

\begin{comment}
我们相信，dentvfm可以提升牙科研究及临床部署效率，并且可以促进人工智能在牙科领域应用的民主化。
使用SSL预训练牙科基础模型需要大量牙科数据，以及大量计算资源，这通常只有发达国家地区的大型专业机构才具备。
我们构建了迄今为止最大的牙科预训练数据集，并在16张h100gpu上进行训练。
我们还提供了多个不同尺寸的dentvfm版本，适合在不同资源限制的场景下使用。
dentvfm被证明具有很高的标签效率，即其在仅有25%标注数据的few-shot设定下仍可实现高性能，在一些场景下可以超过具有50%甚至100%标注数据的基线方法，比如，龋齿评估，骨折诊断，tmj异常诊断，如图a所示。
同时，dengvfm可以作为一个即插即用的模块与参数高效的微调架构集成，实现低计算成本的适配。
通过将dentvfm与vitadapter进行集成，我们可以构建比任务特定模型更好的模型，如图a所示。
dentvfm具有高的标签效率和计算效率，降低了其进一步应用于未来牙科研究与临床应用构建的成本与门槛，对于大多数机构来说更容易实现。
dentvfm促进了人工智能在牙科领域应用的民主化并且降低公众对重复开发任务特定模型带来的对大量资源消耗的质疑。

我们相信dentvfm的开发可以促进人工智能在牙科领域应用的民主化，并提升牙科研究及临床部署效率。
使用SSL预训练牙科基础模型需要大量牙科数据，以及大量计算资源，这通常只有发达国家地区的大型专业机构才具备。
我们构建了迄今为止最大的牙科预训练数据集，并在16张h100gpu上进行训练。
我们还提供了多个不同尺寸的dentvfm版本，适合在不同资源限制的场景下使用。
预训练的dentvfm可以作为一个即插即用的模块，用于替换现有系统中的特征提取模块，并取得更好的表现，如图。
通过与参数高效的微调架构集成，将其微调到特定下游任务的计算要求相对较小，因此对于大多数机构来说更容易实现。
我们的模型可以运行在常用的商用GPU处理器上，如t4、3090、4090等。
另外，在fewshot setting下的性能比较表明，将dentvfm微调到特定任务只需少量的标注数据，如图。
这对进一步降低牙科ai应用的构建成本具有重要意义。
dentvfm对多种牙科任务的通用性可以降低公众对重复开发任务特定模型带来的对大量资源消耗的质疑。
通过开放dentvfm的模型、训练、评测体系，我们期待可以加速ai在牙科的研究和应用。
\end{comment}
DentVFM has the potential to significantly improve the efficiency of dental research and clinical deployment, while also advancing the democratization of AI applications in dentistry. 
Pre-training dental foundation models using SSL requires vast amounts of data and substantial computational resources, which are typically accessible only to large professional institutions in developed regions. 
To overcome this limitation, we construct the largest dental pre-training dataset to date and train DentVFM on 16$\times$NVIDIA H100(80G) GPUs. 
We also offer multiple versions of DentVFM with different model sizes to accommodate varying resource constraints. DentVFM demonstrates remarkable label efficiency, achieving strong performance in few-shot settings with only $25\%$ labeled data. 
It outperforms baselines that rely on data labeled with $50\%$ or even $100\%$, particularly in tasks such as the assessment of caries, the diagnosis of fractures, and the diagnosis of disc displacement of the TMJ, as illustrated in Figure \ref{fig:fewshot_performance}a. 
Furthermore, DentVFM functions as a plug-and-play module that can be seamlessly integrated with parameter-efficient fine-tuning architectures, ensuring low computational costs during adaptation. 
Incorporating DentVFM with ViTAdapter enables the creation of integrated models that outperform task-specific models, as illustrated in Figure \ref{fig:head_to_head_performance}b. 
DentVFM exhibits high label efficiency and computational efficiency, reducing the cost and barrier to further application in future dental research and clinical practice, making it more accessible for most institutions. 
Thus, DentVFM contributes to the democratization of AI applications in dentistry, while alleviating public concerns about the resource consumption associated with the continuous development of task-specific models. 

\begin{comment}
dentvista是一个迄今为止最大的牙科放射图像预训练数据集，包含来自12个地区，7中主要牙科放射图像类型的1.6M牙科放射影像。
dentbench是一个更大、更综合的牙科放射评测基准，包含来自15个地区的覆盖4种主要牙科应用类型、8种牙科专科、超过20种牙科疾病的30个牙科下游任务。

尽管该工作创新性的构建了牙科领域的视觉基础模型，并验证了其在牙科多种应用场景下的优势，但仍然存在一些局限性和挑战，需要在未来的工作中进行探索。
首先，预训练数据主要来源于东亚人群，尽管模型已在更广泛数据上验证，但构建一个涵盖全球多中心、多种族、多疾病谱的放射影像资源库，将是实现真正公平性与普适性的关键。
其次，二维与三维影像的比例失衡限制了体积建模的潜力。未来纳入更多高质量的三维影像，不仅能扩展模型的能力边界，还可能揭示新的解剖学与疾病表征模式。
再次，当前的研究仍局限于视觉模态。牙科诊疗依赖于多维度信息，包括电子病历、影像报告、实验室检查和病理学结果。将这些临床协变量融入预训练框架，不仅可能提升零样本任务的表现，也有望推动 DentVFM 向真正的多模态医学基础模型演化。
此外，与自然语言处理领域的语言模型相比，DentVFM 的参数规模仍相对有限。稳定地训练更大规模的视觉模型，探索在牙科影像学领域是否也存在类似于语言模型中“Scaling Law”的智能涌现，将是未来的关键科学问题。
最后，尽管我们通过 DentBench 显著拓展了牙科任务的覆盖范围，但若要全面评估基础模型的通用智能，还需纳入更具挑战性的任务，如全身性疾病的影像学表现、罕见病诊断以及跨学科口腔—全身健康关联的分析。

综上所述，我们验证了dentvfm在处理牙科放射影像模态多样性方面的有效性，适应多样化牙科疾病和保健应用方面的通用性和高效性，以及在执行替代模态诊断方面表现出的潜力。
摆脱大量高质量标签限制，DentVFM 为牙科 AI 的发展开启了一个新的起点。
通过全球化数据、多模态融合与规模化探索，我们预期未来将出现更具泛化性、更高智能化的牙科基础模型，从而推动口腔医学走向一个更加精准、普惠与智能的新时代。
\end{comment}
Although this work represents an innovative effort to construct a visual foundation model for dentistry, showcasing advantages in diverse dental applications, several limitations and challenges remain, which warrant further exploration in future studies. 
First, pre-training data predominantly consist of samples from East Asian populations. 
Although the model has been validated on diverse tasks, the creation of a radiology image resource database encompassing global, multi-center, multi-ethnic, and multi-disease data will be crucial to achieving true fairness and universality. 
Second, the imbalance between 2D and 3D images limits the potential for volumetric modeling. 
Incorporating a greater number of high-quality 3D images in future iterations will not only enhance capabilities of the model but may also uncover new patterns in anatomical and disease representation. 
Third, the current study is confined to the visual modality. 
Dental diagnosis and treatment depend on multi-dimensional information, including electronic medical records, imaging reports, laboratory test results, and pathology findings. 
Integrating these clinical covariates into pre-training could improve performance on zero-shot tasks and propel DentVFM toward the development of a truly multi-modal medical foundation model. 
Furthermore, compared to language models in natural language processing, the parameter scale of DentVFM remains relatively modest. 
A significant scientific challenge moving forward will be the stable training of larger-scale visual models and exploring whether phenomena akin to the "Scaling Law" in language models also apply to the dental imaging domain, potentially leading to advances in intelligence. 
Finally, while DentBench has considerably expanded the scope of dental tasks, fully assessing the generalist intelligence of the dental foundation model will require the inclusion of more complex tasks, such as rare disease diagnosis, and the analysis of interdisciplinary oral-systemic health correlations. 

In conclusion, we have demonstrated the effectiveness of DentVFM in addressing the diverse dental imaging modalities, showcasing its versatility and efficiency in adapting to a broad spectrum of dental diseases and healthcare applications. 
The model also highlights its potential for performing diagnoses based on surrogate modalities. 
By alleviating the constraints imposed by the requirement for high-quality large-scale annotated data, DentVFM represents a transformative milestone in the evolution of AI for dental research and clinical practice.
Future integration of global data, multimodal paradigms, and large-scale exploration will likely foster dental foundation models with greater generalization and enhanced intelligence, heralding an era of oral medicine distinguished by precision, accessibility, and advanced technological capacity.
\section*{Methods}

\subsection*{Dataset Preparation Process}
\begin{comment}
我们构建了迄今为止最大的牙科多模态无标注放射图像数据集dentvista用于预训练dentvfm。
我们设计了数据预处理pipeline用于标准化不同成像协议下获得的图像以便进行预训练。
为了全面评测各种预训练模型在牙科领域的性能，我们还构建了一个更大的、覆盖更多牙科疾病及下游应用的benchmark dentbench。
下面我们将分别详细说明他们的构建过程。
\end{comment}

We construct DentVista, the largest multimodal unlabeled dental radiographic images dataset to date, for pre-training DentVFM. 
Given the heterogeneous nature of multimodal data, we design a data preprocessing pipeline to standardize images acquired under different imaging protocols for pre-training. 
To comprehensively evaluate the performance of various pre-trained models, we construct DentBench, a larger benchmark that encompasses a broader spectrum of dental diseases and downstream applications. 
The data collection and pre-processing pipelines for DentVista and DentBench are illustrated in Figure \ref{fig:dataset_statistic}b. 
We will elaborate on their respective construction processes below. 

\subsubsection*{Curation of DentVista}
\begin{comment}
dentvista数据的收集和预处理pipeline见图1b所示。
dentvista中的牙科多模态放射数据来源于一家顶尖的医院（九院）、105家牙科诊所以及网络上可以公开获取的牙科预训练数据集覆盖全球12个地区的多个医疗中心。
我们从合作机构（1家医院和105家诊所）的电子档案中提取来自2020~2024年间的放射影像数据，其构成了dentvista的主体。
另外，我们还纳入少量网络公开可获取的无标注牙科影像数据作为补充。
为了防止数据泄露以及评测模型对分布外数据的泛化能力，我们将大部分牙科公开数据集作为外部评测集，仅纳入少量包含无标注数据的数据集进入预训练。
关于数据来源的详细说明可以参考表1。
最终，dentvista包含了约1.6M的放射影像（大约30M的slices）覆盖7中主要的牙科放射学影像。
关于dentvista的详细统计信息可以参考section 1。
来源于不同设备、不同模态的图像遵循多种成像protocol。
为了将这些数据用于预训练，我们构建了数据预处理pipeline，用于数据标准化。
具体来说，我们根据隐私保护的要求，首先对所有数据进行匿名化，以去除身份标识。
然后，我们过滤掉低质量图像数据，并将像素归一化到0~255范围内。
对于3D体数据（CT、CBCT、MRI），我们分别从失状向，冠状向，轴向随机提取2dslices。
我们获得了约3M的2d影像用于dentvfm2d的预训练，约311k volumes用于dentvfm3d的预训练。
\end{comment}

The multimodal dental radiographic data in DentVista originate from 3 Chinese hospitals, 105 dental clinics, and some publicly available data from the Web, covering multiple medical centers in 12 global regions.  
We extract imaging records of patients who were seen in collaborating institutions (3 hospitals and 105 clinics) between 2020 and 2024. 
In addition, we incorporate a small amount of publicly available unlabeled datasets from the Web as complementary data. 
Here, existing public labeled datasets are used as external evaluation sets to prevent data leakage during pre-training and evaluate the model's generalization capability under out-of-distribution settings. 
More detailed information on data sources can be found in the Supplementary Table \ref{tab:dentvista_sources}. 
Ultimately, DentVista comprises approximately 1.6M multimodal radiographic images (around 30M slices) covering 7 major types of dental radiological imaging. 
Detailed statistics are provided in the \hyperref[sec:result]{Results} section. 
Moreover, images from different devices follow diverse imaging protocols. 
To ensure consistent input for pre-training, we construct a data preprocessing pipeline for data standardization (refer to Figure \ref{fig:dataset_statistic}b). 
Specifically, we initially perform data anonymization to remove identification following the privacy protection policy. 
Then, we filter out low-quality images based on image statistical features (e.g., signal-to-noise ratio, information entropy, grayscale histogram) and perform normalization. 
Specifically, the pixel values of the 2D radiographic images are normalized to the range of 0 to 255. 
The Hounsfield Unit (HU) intensities of the volumetric data (CT, CBCT) and the signal intensity of MRI are scaled to the range of 0 to 1. 
This normalization is based on the 0.5\% and 99.5\% percentiles, with intensity values outside this range being clipped. 
We crop the foreground regions from all images. 
For volumetric data, we randomly extract 2D slices from the sagittal, coronal, and axial planes to pre-train DentVFM-2D. 
Finally, we obtained approximately 3M 2D images for DentVFM-2D pre-training and about 311K volumes for DentVFM-3D pre-training.

\subsubsection*{Curation of DentBench}
\begin{comment}
Dentbench是一个更大、更全面的牙科放射评测基准，纳入了21个公开的牙科任务作为外部评测数据以及18个精心构建的任务作为内部评测数据。
我们在result中说明了dentbench的详细统计信息。
对于每个任务的详细描述及任务来源如表1所示。
现存的开源数据集包含了有限的牙科疾病，有限的影像类型并且主要是诊断和分割类任务，和治疗相关的任务稀少。
为此，我们精心构建了内部任务作为补充以扩充疾病数量（e.g. cyst、tmj、牙周炎、骨质疏松）、任务类型（e.g. 治疗分析以及生物标志物识别）、影像类型（e.g. MRI）
这些内部评测数据是回顾性的，来源于在九院就诊的患者。
为了防止预训练中的数据泄露，我们将被用于内部评测数据的患者的所有影像（根据放射编号进行关联）从dentvista中彻底移除。
所有数据的标签信息都来自于患者病历中的诊断，并经过多名有经验的牙科医生的手动检查校正。
我们对数据执行匿名化以符合隐私保护协议。
\end{comment}
DentBench is a more extensive and comprehensive dental radiographic evaluation benchmark, incorporating 22 publicly available dental datasets on the Web as external evaluation tasks in addition to 16 meticulously constructed datasets that serve as internal evaluation tasks. 
Detailed statistics of DentBench is presented in the \hyperref[sec:result]{Results} section, with detailed task descriptions and sources provided in the Extended Data Table \ref{tab:dataset-summary}. 
To the best of our knowledge, existing public datasets\cite{wang2016benchmark,panetta2021tufts,silva2023boosting,li2024multi} cover a limited range of dental diseases and imaging modalities, mainly focused on diagnostic and segmentation tasks with few treatment-related applications. 
To address these limitations, we carefully develop some internal tasks to increase the coverage of the disease (e.g. cysts, temporomandibular joint disorders, periodontitis, osteoporosis), expand task categories (e.g., treatment analysis and biomarker identification), and diversify imaging types (e.g., MRI). 
These internal tasks comprise retrospective data derived from patients treated at Shanghai Ninth People's Hospital. 
To prevent data leakage during pre-training, we systematically remove all images (cross-referenced by radiographic identification numbers) from patients included in internal evaluation datasets from DentVista. 
All data annotations are extracted from medical records and have undergone rigorous manual verification by multiple experienced dental clinicians. 
Data anonymization is implemented to ensure compliance with privacy protection standards.

% TODO：此处应该添加说明将下游任务按照7：3划分为训练集和测试集

\subsection*{Large-scale Visual Pre-training}
\begin{comment}
dentvfm通过大规模视觉预训练获得牙科通用智能。
模型架构和预训练算法在视觉预训练中扮演重要角色。
对于在dentvista上的大规模预训练，我们选择vit作为基础架构，并使用dinov2，一个sota自监督学习方法基于自蒸馏技术进行预训练。
下面我们将对基础模型架构和预训练protocol进行详细说明。
\end{comment} 
DentVFM acquires the dental generalist intelligence through large-scale visual pre-training. 
Both the model architecture and the pre-training algorithm play critical roles in visual representation learning.
For large-scale pre-training on the DentVista dataset, we adopt Vision Transformer (ViT)\cite{dosovitskiy2020image} as the backbone architecture and employ DINOv2\cite{oquab2023dinov2}, a state-of-the-art self-supervised learning method based on self-distillation.
In the following sections, we provide a detailed description of the model architecture and the pre-training protocol.

\subsubsection*{Backbone architecture}
\begin{comment}
我们选择vanilla vit模型作为dentvfm的基础架构。
images首先被划分为多个patch（2d patch for dentvfm-2d，3d patch for dentvfm-3d），然后经过线性映射获得patch embedding。
position embedding被添加进patch embedding中，同时，一个称为cls的token被附加到所有token中。
Transformer被用于建模token之间的依赖关系。
尽管一些工作通过为transformer添加局部空间运算，将特定于视觉的归纳偏置引入，提升了transformer在dense prediction任务上的效果，但是，vanilla vit依然有一些不可忽视的优势，
vanilla vit对于预训练更加灵活（比如进行MIM），同时，它具有很强的扩展性，它可以方便的与其他现有模型集成（比如各种先进的adapter架构以及LLM）。
我们对不同的vit变体进行了预训练，包括vitb、vitl、vitg，为不同的资源-performance平衡场景提供开箱即用的模型。
\end{comment} 
We adopt the vanilla Vision Transformer (ViT)\cite{dosovitskiy2020image} as the backbone architecture for DentVFM. 
Input images are first partitioned into patches sequences, 2D patches for DentVFM-2D and 3D patches for DentVFM-3D, which are then linearly projected to generate patch tokens. 
The resolutions of the patches are $14\times14$ for 2D patches and $16\times16\times16$ for 3D patches.  
To retain spatial information, positional embeddings are added to the patch tokens. 
A learnable [CLS] token is inserted into the token sequence. 
A standard transformer is used to model the dependencies among all tokens.
Although several studies have enhanced transformer performance for dense prediction tasks by introducing vision-specific inductive biases into model architectures\cite{liu2021swin,liu2022swin,wang2021pyramid}, the vanilla ViT retains distinct advantages. 
Its architecture remains highly adaptable for pretraining objectives such as masked image modeling (MIM) and exhibits excellent scalability, enabling seamless integration with other advanced models such as adapter modules and large language models (LLMs). 
To balance computational efficiency and effectiveness, we provide multiple pre-trained ViT variants (i.e. ViT-B, ViT-L, and ViT-G) offering flexible, plug-and-play solutions for a variety of resource-constrained deployment scenarios. 
The architectural details of these variants can be found in the Supplementary Table \ref{tab:model_architecture}.

\subsubsection*{Pre-training protocol}
\begin{comment}
我们使用最近被提出的先进自监督预训练框架（dinov2）进行预训练。
dinov2是一个判别式自监督方法改进自dino和ibot。
dinov2遵循一个知识蒸馏范式where we训练一个student network去匹配teacher network的输出。
teacher网络的参数通过EMA方法进行更新。
student网络的预训练代理任务包含了来自dino的image-level objective以及来自ibot的patch-level objective。

image-level objective是由来自teacher和student模型的cls特征计算的CEloss。
具体来说，从一张给定的image，我们构造a set of different distored views using multi-crop strategy。
\end{comment} 

We adopt DINOv2\cite{oquab2023dinov2}, a recently proposed state-of-the-art self-supervised pre-training framework. 
DINOv2 is a discriminative self-supervised learning method that extends DINO\cite{caron2021emerging} and iBOT\cite{zhou2021ibot}. 
It follows a knowledge distillation paradigm, in which a student network $g_{\theta_s}$ is trained to match its output with that of a teacher network $g_{\theta_t}$. 
The optimization objective of the student network combines an image-level objective, inherited from DINO, with a patch-level objective, derived from iBOT. 
The teacher and student networks share the same architecture which consists of a backbone, a DINO head for image-level objective computation, and an iBOT head for patch-level objective computation. 
Here, the DINO head and the iBOT head are two separate MLPs. 
An exponential moving average (EMA)\cite{he2020momentum} is used to update the weights of the teacher network, i.e. $\theta_t \leftarrow \lambda \theta_t + (1 - \lambda) \theta_s$.
The overview of the pre-training protocol is shown in Figure \ref{fig:main_framework}c.

The image-level objective is the cross-entropy loss between the image-level features extracted from the student and the teacher network. 
Both image-level features come from the [CLS] tokens of backbones, obtained from different views of the same image. 
More precisely, a set of different distorted views, $V$, is generated from a given image $x$ based on a multi-crop strategy\cite{caron2020unsupervised}. 
This set contains two global views, $x_{1}^{g}$ and $x_{2}^{g}$, and several local views. 
The resolutions of the global views are $224\times224$ for DentVFM-2D and $96\times96$ for DentVFM-3D. 
In contrast, the resolutions of the local views are set $98\times98$ for DentVFM-2D and $48\times48$ for DentVFM-3D. 
All views are passed through the student network, while only global views are passed through the teacher network. 
For each view, we obtain the [CLS] tokens of the backbones. 
We pass the student [CLS] tokens through the DINO head of the student network and apply a softmax to obtain the probability distributions $P_{s}^{d}(x)$. 
Similarly, we pass the teacher [CLS] tokens through the DINO head of the teacher network and apply a softmax followed by a Sinkhorn-Knopp centering\cite{ruan2022weighted} to obtain the probability distributions $P_{t}^{d}(x)$. 
The image-level objective corresponds to the following:
\[
\mathcal{L}_{\text{image-level}} =  
\min_{\theta_s} \sum_{x \in \{x_1^g, x_2^g\}} \sum_{\substack{x' \in V \\ x' \neq x}} H\big(P_t^{d}(x), P_s^{d}(x')\big)
\],
where $H(a,b) = -a \log b$.

The patch-level objective is the cross-entropy loss between visible patch tokens from the teacher and corresponding masked patch tokens from the student. 
Specifically, we perform blockwise masking\cite{bao2021beit} on a view, e.g. $x$, and obtain a masked view $\hat{x}$. 
The masked view $\hat{x}$ is passed through the student network, and the original view $x$ is passed through the teacher network. 
The masked tokens from the student backbone are fed to the student iBOT head and then applied softmax to obtain the probability distributions, for example, $P_{sj}^{i}(\hat{x})$ for the masked token $j$. 
Similarly, we apply the teacher iBOT head to the visible tokens from the teacher backbone and then use softmax and centering steps to obtain the probability distributions, e.g. $P_{tj}^{i}(x)$ for the corresponding token $j$. 
The patch-level objective corresponds to the following:
\[
\mathcal{L}_{\text{patch-level}} =  
\min_{\theta_s} \sum_{j}H\big(P_{tj}^{i}(x), P_{sj}^{i}(\hat{x})\big)
\],
where $j$ are patch indices for masked tokens.

DINOv2 is directly applicable to 2D dental radiographic images. 
However, its view-augmentation pipeline is not inherently suited for volumetric data. 
To address this limitation, we design a custom view augmentation pipeline tailored for pretraining on volumetric data. 
Specifically, we replace the standard 2D image cropping operation with its 3D counterpart. 
In addition, we substitute the typical brightness, contrast, and saturation adjustments used for natural images with contrast enhancement techniques specifically optimized for medical images. 
Furthermore, the horizontal flip operation is replaced by flips along all three spatial axes to account for the volumetric nature of the data.  
We choose Adam\cite{kingma2014adam} as the optimizer. 
A warm-up phase is applied. 
More details on hyperparameters (e.g. batch size, learning rate, weight decay, iteration number) are provided in the Supplementary Table \ref{tab:hyperparameters}.

\subsection*{Evaluation Framework}
\begin{comment}
为了全面评估预训练模型在不同牙科应用场景中的表现，我们设计了一个多维度评估框架。
我们选择了多种预训练模型作为baselines用于对比，它们覆盖多种预训练算法、多种预训练数据。
为了直接比较不同预训练模型提取通用表示的能力，我们在冻结的预训练模型上附加轻量化的任务特定的分类头和分割头并在各种下游任务上进行finetune。
为了评测模型在资源受限条件下的性能，我们在finetune的过程中使用具有不同训练数据量的设定。
为了评估模型的可扩展性和即插即用的能力，我们将预训练的模型作为一个替换模块同vitadapter集成起来。
为了评估模型的替代模态诊断能力，我们在专门设计的，具有替代模态诊断特点的数据集上对dentvfm的adapter集成进行评测。
另外，我们通过应用不同设置，包括不同预训练模型参数量、不同预训练数据、不同预训练算法对模型进行全面的消融分析。
每种评估维度的详细设置将在下面进行说明。
\end{comment} 

To comprehensively evaluate the performance of pre-trained models across various dental applications, we design a multi-dimensional evaluation framework. 
In the comparisons, we select a set of existing pre-trained models and task-specific models as baselines.
The evaluation framework encompasses assessments of the dental generalist intelligence of pre-trained models, the performance under few-shot learning settings, the plug-and-play compatibility, the surrogate modality diagnostic capability, and an ablation analysis of key factors of pre-training. 
Detailed configurations for each evaluation will be presented in the following sections.

\subsubsection*{Comparisons and baselines}

For the dental generalist intelligence evaluation, we compare DentVFM against 11 pre-trained models commonly used in the medical imaging analysis community. 
These pre-trained models can be categorized according to the pretraining algorithm, model architecture, and the type of pretraining data. 
In terms of the pre-training algorithm, they are divided into supervised (Resnet50\cite{he2016deep}, SAM\cite{kirillov2023segment}, SAM\_Med2d\cite{cheng2023sammed2d}, SAM\_Med3d\cite{wang2024sam}, SwimUNETR\cite{tang2022self}), weakly supervised (CLIP\cite{radford2021learning}, BiomedCLIP\cite{zhang2023biomedclip}, M3D\cite{bai2024m3d}), and self-supervised pre-training (DINOv2\cite{oquab2023dinov2}, LVM-Resnet50\cite{mh2023lvm}, LVM-ViT\cite{mh2023lvm}). 
With respect to the model architecture, they are classified into Resnet-based (Resnet50, LVM-Resnet50), ViT-based (CLIP, SAM, DINOv2, BiomedCLIP, SAM\_Med2d, SAM\_Med3d, LVM-ViT, M3D) and Swim-Transformer-based (SwimUNETR) frameworks. 
Regarding the pre-training data, these models are distinguished by the use of natural image datasets (Resnet50, CLIP, SAM, DINOv2) or medical image datasets (other baselines). 
More details on baselines can be found in Supplementary Table \ref{tab:baseline_details}.
In our implementation of these pre-trained models, we use their official model checkpoints. 
Here, we select ViT-B for pre-trained models (i.e. CLIP, SAM, DINOv2) that provide multiple checkpoint versions. 
For comparisons in few-shot settings, we select the models that performed well in the generalist intelligence evaluation for the corresponding tasks as baselines. 
For the evaluation of plug-and-play compatibility, we select several task-specific methods for comparison to demonstrate that DentVFM, when integrated with advanced adapter frameworks, can outperform task-specific models. 
Specifically, for the dental cyst diagnosis task (\textit{FG Cyst Diag}), we select LCD-Net\cite{hu2021location} as a baseline. 
For the TMJ abnormality diagnosis task (\textit{TMJ Abnl Diag (PAN)}), a fully fine-tuned Resnet50\cite{he2016deep} is chosen as the baseline. 
For the dental caries segmentation task (\textit{Caries Seg}), UNet\cite{ronneberger2015u} and MLUA\cite{wang2023multi} are used as baseline methods. 
For both the apical periodontitis segmentation task (\textit{Pal Seg}) and the oral structure segmentation task (\textit{Oral Struct Seg (TF3)}), we select the representative and robust model, nnUNet\cite{isensee2021nnu,isensee2024nnu}, for comparison. 
The reproduction of these task-specific models follows their default settings.  
In ablation experiments, we choose MAE\cite{he2022masked}, a self-supervised algorithm commonly used in medical image pre-training, for comparison. 
The settings of MAE follow its default configuration. 

\subsubsection*{Generalist intelligence evaluation settings}
\begin{comment}
为了直接比较不同的预训练模型提取的表示的通用性，我们在冻结的预训练模型上添加可训练的轻量化的任务特定的分类模块或分割模块用于dentbench中各种任务的评测。
对于分类任务，我们对预训练的模型执行linear probing。
具体来说，我们先使用预训练模型提取一个任务训练集中图像的图像级别表示，然后用这些图像级别的表示及图像类别标签训练一个用于该分类任务的逻辑回归器。
对于vit模型，图像级别的表示从最后一层的cls token获得，对于resnet模型，图像级别的表示则是图像特征经过全局池化后的得到的特征。
在训练逻辑回归模型时，我们对正则化强度倒数c进行超参数搜索以平衡偏差和方差，从而选择近似的最佳模型。
这里，我们采样了45个超参数c，其值从10-6~105。
对于分割任务，我们为2d图像和3d图像应用了不同的轻量化分割头。
对于2d图像，我们在预训练模型上添加一个简单的线性分割头with batch normalization，该分割头的输入为预训练模型4层中间图像表示上采样到输入图像分辨率后的拼接。
对于3d图像，我们应用一个来自unetr的分割头。
此处展开描述unetr如何做的。
对于分割头的预训练，我们使用adam优化器以及0.0001的初始学习率。
为了稳定的评估不同模型的效果，我们对每个任务的数据集进行5次随机train-test划分，并统计评测指标的均值和标准差。
\end{comment}

To directly compare representations extracted by different pre-trained models, we append lightweight task-specific classification or segmentation modules to the pre-trained models for evaluation on various tasks in DentBench. 
During fine-tuning for downstream tasks, we keep the weights of the pre-trained model frozen and only update the weights of the attached modules. 
This setup minimizes the influence of other factors, allowing a direct comparison of the generalization of representations extracted by different pre-trained models. 
To ensure stable evaluation across the entire task dataset, we perform 5 random train-test splits for each dataset and report the mean and standard deviation of the evaluation metrics. 

For classification tasks, we perform the linear probing. 
Specifically, we first use the pre-trained model to extract image-level representations from the training set of a task, and then train a logistic regressor on these representations and corresponding labels. 
The image-level representation is derived from the [CLS] token of the last layer in ViT backbones and from the globally average pooled image features in ResNet backbones. 
During training of the logistic regression model, we perform a hyperparameter search over the inverse regularization strength, $C$, to balance bias and variance. 
A total of 45 $C$ values are sampled on a logarithmic scale from $10^{-6}$ to $10^{5}$. 
The best $C$ based on the validation performance is then used to evaluate on the test set. 
Optimization of the logistic regression model is allowed up to 1000 iterations, with the stopping criterion set to a tolerance of $10^{-12}$. 

For segmentation tasks, we use different lightweight segmentation modules for 2D and 3D radiologic images. 
A simple linear segmentation head with batch normalization is added on top of the frozen pre-trained model for 2D images. 
The segmentation head takes as input the concatenated, interpolated representations from the last four layers of the pre-trained model, aligned to the resolution of the input image. 
Let the input image be $x\in \mathbb{R}^{W\times H}$, and the output of the feature map of the $i$-th layer can be denoted by $z_i\in \mathbb{R}^{W_i\times H_i \times K}$, where $W_i$ and $H_i$ are the shape of the feature map and $K$ is the dimension of the feature. 
$z_i$ will be first interpolated to $\hat{z_i}\in \mathbb{R}^{W\times H\times K}$. 
Then, all interpolated representations, $\hat{z_i}(i\in T)$, from the set of target layers $T$ are concatenated as the input of the linear segmentation head. 
We used the UNETR\cite{hatamizadeh2022unetr} segmentation head for 3D radiographic images. 
UNETR integrates the ViT encoder into the UNet\cite{ronneberger2015u} framework, where features from multiple resolutions of the encoder are combined with the decoder. 
In the implementation of the UNETR segmentation head, we first extract a set of patch tokens $Z=\{ z_i\in \mathbb{R}^{\frac{W}{P}\times \frac{H}{P}\times \frac{D}{P}\times K} | i\in T \}$ from the ViT backbone, where $T=\{(1+j)\frac{L}{4} | j\in\{0,1,2,3\} \}$ and $L$ is the layer number of a certain ViT version, then reshape and project them into different input spaces of different resolutions utilizing consecutive convolutional and deconvolutional operations. 
More details about UNETR head can refer to the default settings of UNETR. 
In the fine-tuning of segmentation modules, we use the Adam\cite{kingma2014adam} optimizer with an initial learning rate of 0.0001. 
We train for 300 epochs with a batch size of 32.

\subsubsection*{Few-shot evaluation settings}
\begin{comment}
为了评估模型在资源受限场景下的性能，我们从一个目标任务的训练集中采样少量的数据模拟有限标注的场景。
我们首先随机对任务的数据进行train-test划分。
然后，我们对得到的训练集按照比例K%进行随机采样。
这里，采样比例K取25%、50%、75%，100%。
其中，100%表示使用整个训练集进行训练。
对于不同的K，测试集保持不变。
考虑到采样的训练数据不同会对结果产生很大影响，为了降低偏差，我们对每个K的取值进行了5次随机采样，并重新finetune对应模型。
对每个K，我们计算5次随机采样后训练得到模型的均值和标准差。
在few-shot评测中，我们使用的模型架构和通用智能测试中使用的相同。
\end{comment} 

To evaluate performance under scare-label conditions, we simulate limited annotations by sampling a small subset of training data from a downstream target task. 
First, we perform a random train-test split on the target task dataset. 
Then, a proportion $k\%(25\%, 50\%, 75\%, 100\%)$ is randomly sampled from the training set, where $100\%$ indicates using the complete training set. 
For each value of $k$, we use the same testing set. 
We use the same model architectures and fine-tuning configurations as in the generalist intelligence evaluation during the fine-tuning of the few-shot evaluation. 
Here, the downstream tasks we selected include classification and segmentation for both 2D and 3D radiologic images. 
Random sampling of training data can have a significant impact on fine-tuning. 
Therefore, we perform 5 random samplings for each $k$ and fine-tuning the corresponding model to reduce variance. 
We compute the mean and standard deviation of the performance metrics from the five random samplings for each value $k$.

\subsubsection*{Compatibility evaluation settings}
\begin{comment}
当前，已经存在许多工作致力于将vit应用到各种视觉下游任务中，并取得了令人印象深刻的效果。
dentvfm可以作为一个即插即用的模块与这些先进的方法进行集成。
由于预训练的vit通常包含大量的参数，对其进行全量微调需要大量的数据。
为了高效地将预训练vit适配到下游任务上，许多高效的微调架构被开发出来。
dentvfm可以与这些现有的微调框架兼容。

对于分类任务，我们将dentvfm作为线性adaptation的一部分，即，在冻结的dentvfm后添加一个可训练linear layer用于分类。
linear layer的输入为最后一层transformer得到的cls token，或者是多层的cls token的average。
我们选择了两个有代表性的任务，a和b，进行评测。
在finetune的过程中，我们选择adam优化器，并对learning rate和特征层数进行超参数grid search。
learning rate的取值为[1e-5, 2e-5, 5e-5, 1e-4, 2e-4, 5e-4, 1e-3, 2e-3, 5e-3, 1e-2, 2e-2, 5e-2, 0.1]，特征层数的取值为【1，4】。
我们对训练数据应用水平flip进行增强。
我们总共训练12500次迭代，并报告最佳结果。
我们对任务数据集执行了5次随机train-test划分，并对每个划分执行finetune和评测。
对于分割任务，我们对2dimage和3dimage应用不同的分割框架。
具体来说，我们将dentvfm2d和mask2former的分割头相结合用于2d image分割，将dentvfm3d和unetr的分割头相结合，用于3d image的分割。
另外，我们对预训练的vit backbone应用vitadapter。
此处添加对vitadapter的详细描述。
\end{comment} 

Recently, many works\cite{cheng2021per,cheng2022masked,carion2020end} have focused on applying ViT to various visual tasks, achieving impressive results. 
DentVFM can be seamlessly integrated as a plug-and-play module with these advanced methods. 
Since pre-trained ViT models typically contain a large number of parameters, fine-tuning them requires substantial amounts of data. 
To efficiently adapt pre-trained ViTs to downstream tasks, several efficient fine-tuning frameworks have been developed. 
DentVFM can be compatible with these advanced adaptation frameworks. 
We apply different existing frameworks for classification and segmentation tasks to demonstrate the compatibility of DentVFM. 

For classification tasks, we incorporate DentVFM into a linear adaptation framework. 
Specifically, a learnable linear layer is added after the frozen DentVFM for classification. 
The input to the linear layer is either the [CLS] token from the last layer of the backbone or the average of the [CLS] tokens from multiple layers. 
We select two representative tasks, \textit{FG Cyst Diag} and \textit{TMJ Abnl Diag (PAN)}, for evaluation. 
During fine-tuning, we use the Adam optimizer and conduct a grid search over the learning rate and the number of feature layers as hyperparameters. 
The learning rates are chosen from the set $\{1\mathrm{e}{-5}, 2\mathrm{e}{-5}, 5\mathrm{e}{-5}, 1\mathrm{e}{-4}, 2\mathrm{e}{-4}, 5\mathrm{e}{-4}, 1\mathrm{e}{-3}, 2\mathrm{e}{-3}, 5\mathrm{e}{-3}, 1\mathrm{e}{-2}, 2\mathrm{e}{-2}, 5\mathrm{e}{-2}, 0.1\}$ and the number of feature layers are selected from $\{1,4\}$. 
We apply horizontal flipping enhancement to the training data and train the models for a total of 12,500 iterations, reporting the best results. 
We perform 5 random train-test splits and conduct fine-tuning and evaluation for each split. 

For segmentation tasks, we apply different segmentation frameworks for 2D and 3D images. 
Specifically, we combine DentVFM-2D with the Mask2Former\cite{cheng2022masked} segmentation head for 2D image segmentation and DentVFM-3D with the UNETR segmentation head for 3D image segmentation. 
Additionally, we integrate pre-trained models with ViTAdapter\cite{chen2022vision}. 
ViTAdapter improves the performance of dense prediction tasks by introducing image-based inductive biases into the vanilla ViT architecture. 
ViTAdapter designs a spatial prior module (SPM) to model the local spatial context based on convolutions. 
Following the default configuration, we use a stack of stride-2 $3\times3$ convolutions to obtain a feature pyramid $\{F_1, F_2, F_3 \}$, which contains $K$-dimensional feature maps with resolutions of $\frac{1}{8}$, $\frac{1}{16}$, and $\frac{1}{32}$. 
Here, we set $K$ to the same as the dimension of the hidden feature of the corresponding ViT. 
The feature maps of the feature pyramid are flattened and concatenated into feature tokens denoted by $F_{sp}^{1}\in \mathbb{R}^{(\frac{HW}{8^2}+\frac{HW}{16^2}+\frac{HW}{32^2})\times K}$. 
ViTAdapter uses two feature interaction modules, called the Spatial Feature Injector and Multi-Scale Feature Extractor, to bridge the feature maps of SPM and ViT. 
Both modules are mainly based on the cross-attention mechanism\cite{vaswani2017attention}. 
For the Spatial Feature Injector module, we take the feature $F_{vit}^{i}$ from the $i$-th layer of the ViT backbone as the query, and the spatial feature $F_{sp}^{i}$ as the key and value. 
The update process of $F_{vit}^{i}$ can be written as:
\[
\hat{F_{vit}^{i}} =  F_{vit}^{i} + \gamma^{i}Attention(norm(F_{vit}^{i}), norm(F_{sp}^{i}))
\],
where $norm(\cdot)$ is LayerNorm\cite{ba2016layer}. 
For the Multi-Scale Feature Extractor module, another cross-attention layer and a feed-forward network (FFN) are used to update the spatial feature. 
This process can be formulated as follows.
\[
F_{sp}^{i+1} =  \hat{F_{sp}^{i}} + FFN(norm(\hat{F_{sp}^{i}})),
\]
\[
\hat{F_{sp}^{i}} =  F_{sp}^{i} + Attention( norm(F_{sp}^{i}), norm(F_{vit}^{i+1}))
\], 
where $F_{sp}^{i}$ is the query and $F_{vit}^{i+1}$ is used as the key and value. 
We customize a new SPM based on 3D convolutions to adapt DentVFM-3D. 
Therefore, the flattened spatial feature tokens are $F_{sp}^{1}\in \mathbb{R}^{(\frac{HWD}{8^3}+\frac{HWD}{16^3}+\frac{HWD}{32^3})\times K}$. 
More configurations follow the default settings of ViTAdapter\cite{chen2022vision}.

\subsubsection*{Corss-modality diagnosis evaluation settings}
\begin{comment}
我们选择cyst和tmj任务用于替代模态诊断评测。
cyst诊断任务的目标是从全景xray中直接区分口腔囊肿的类型（成釉细胞瘤、角化囊肿、含牙囊肿、根尖囊肿）。
通常来说，对口腔囊肿的诊断需要借助病理分析。
TMJ诊断任务则是根据全景xray判断患者是否存在髁突关节盘异常，这种异常通常需要根据TMJ部位的MRI进一步分析。
我们选择的任务代表了一种使用替代模态数据进行诊断的场景，即使用部分模态数据诊断原本需要更多模态数据才可以进行诊断的疾病。
替代模态的诊断能力对医疗资源有限的地区和医疗机构有重要意义。
在评测中，我们使用linear adaptation框架，类似Compatibility evaluation setting中的描述。
我们将任务特定的专用模型以及具有不同经验的临床医生的手动评估作为baselines。
对于临床医生的评测，我们邀请了6名具有不同经验的临床医生。
3年以下经验的医生称为初级队列，3-8年称为中级队列。
我们要求医生手动对测试集中的数据进行类别诊断。
使用共识诊断，超过一半的医生诊断为同一个类别即确定该类别，如果发现不一致，则讨论达成一致后确定诊断类别。
我们对5次随机train-test划分得到的5个test数据集进行手动评测，并汇报其评测结果。
\end{comment} 

We select \textit{FG Cyst Diag} and \textit{TMJ Abnl Diag (PAN)} tasks to evaluate the diagnosis based on surrogate modality. 
The goal of the cyst diagnosis task is to differentiate between four types of oral cysts (ameloblastoma, dentigerous cyst, keratocyst, and periapical cyst) using panoramic X-rays. 
Typically, a detailed diagnosis of oral cysts requires the support of a pathological analysis. 
The TMJ abnormity diagnosis task involves determining whether a patient has abnormalities in the condyle and joint disk based on panoramic X-rays, which usually necessitates further investigation through MRI of the TMJ region. 
These tasks represent scenarios in which alternative modality data are employed for diagnosis, i.e., diagnosing conditions that typically require multiple modalities using data from only one modality. 
This approach is particularly valuable in regions and medical institutions with limited healthcare resources. 
In the evaluation, we employ a linear adaptation framework described in the compatibility evaluation section, as well as the same configurations. 
We compare the performance of DentVFM with some task-specific models (i.e. LCD-Net\cite{hu2021location} and Resnet50\cite{he2016deep}) and manual evaluations conducted by three dental clinicians with at least five years of clinical experience as baselines. 
All clinicians are recruited from the Shanghai Ninth People's Hospital. 
Each clinician is required to independently diagnose all samples in the test set. 
The final predicted category for each sample is determined based on the consensus of all clinicians. 
Specifically, for each sample, if more than half of the clinicians selected the same diagnosis category, that category is chosen as the predicted label. 
Otherwise, experts will discuss and reach a consensus on the diagnostic category. 
Manual evaluations are conducted on five test datasets, obtained from five random train-test splits, and the evaluation results are reported.

\subsubsection*{Ablation analysis settings}
\begin{comment}
我们在消融实验中研究模型大小、预训练数据大小、预训练数据类别、预训练数据算法对DentVFM的影响。
对于模型大小、预训练数据集大小的影响分析，我们选择了两个有代表性的任务进行，即a和b。
我们独立训练了不同大小的vit版本（包括vitb、vitl、vitg），模型的详细架构可以参考table1。
我们从dentvista中随机采样包含10k图像的子集作为小数据集预训练的设置。
我们对每个任务对应的数据集都执行了5次随机train-test划分，并汇报平均结果。
对于不同预训练算法的影响分析，我们使用常被用于医学图像预训练的MAE算法在我们的dentvista上进行预训练。
这里我们使用vit-b作为backbone。
我们将预训练好的模型在dentbench的所有分类任务上进行评测以评估其综合性能。
对于不同预训练数据类别的影响分析，我们从dentvista中提取所有全景xray作为一个单一模态的预训练子集，并对一个vitb架构的backbone进行预训练。
我们选择了一些有代表性的分类和分割任务进行评测。
分类任务包含abc。
分割任务包含cde。
这些任务都是比较困难的，需要根据图像的细节进行分析。
\end{comment} 

In the ablation study, we investigate the impact of model size, pre-training dataset size, pre-training data categories, and pre-training algorithms on DentVFM. 
To analyze the effects of model size and pre-training dataset size, we select two representative tasks, \textit{Oral Abnl Diag (DENTEX)} and \textit{FG Cyst Diag}. 
We independently train different versions of the ViT backbones (including ViT-B, ViT-L, and ViT-G), with the detailed information on these versions provided in the Supplementary Table \ref{tab:model_architecture}. 
We randomly sample a subset of 10k images from DentVista to construct a small pre-training dataset. 
We use the same evaluation architecture in the generalist intelligence evaluation and perform five random train-test splits. 
For the analysis of the impact of different pre-training algorithms, we use the MAE\cite{he2022masked} algorithm, a method commonly used for medical image pretraining, to pre-trained a new ViT-B on our DentVista dataset. 
We evaluate the MAE pre-trained model on all classification tasks in DentBench to assess its overall performance. 
To investigate the impact of different pre-training data categories, we extract all panoramic X-rays from DentVista to create a single-modality pre-training subset and be used to pre-train a new ViT-B backbone. 
Several representative classification and segmentation tasks are selected for evaluation. 
The classification tasks include \textit{FG Perio Grading}, \textit{Caries Assess}, and \textit{POI Interp (BiMax)}, while a segmentation task \textit{Caries Seg}. 
These tasks are challenging and require detailed analysis of the image content.

\subsection*{Model Visualization Method}
\begin{comment}
为了直观的解释模型的表示学习能力，我们对表示进行了多粒度的可视化。
我们使用tsne对image-level表示进行可视化。
同时，我们还可视化了MHSA用于展示image-level表示与图像不同区域的关联。
我们使用无监督聚类算法k-means可视化pixel-level的表示。
对于体素级别的表示，我们同样使用tsne可视化每个volume表示的分布。
\end{comment} 
To intuitively demonstrate the representation learning capabilities of pre-trained models, we perform multi-granularity visualizations of learned features across three levels: image, pixel, and voxel. 
At the image level, we employ t-SNE\cite{maaten2008visualizing} to reduce dimensions and visualize image-level representations. 
Furthermore, we visualize multi-head self-attention (MHSA) maps to elucidate the associations between image-level representations and different anatomical regions. 
At the pixel level, we apply the k-means clustering algorithm to group and visualize pixel-level representations. 
At the voxel level, we again employ t-SNE to reduce dimensions and visualize voxel-level representations of 3D volumes.

\subsubsection*{Visualization of image-level representations}
\begin{comment}
我们将dentvfm的cls token作为image-level的表示。
我们选择了两个数据集（age,abnl）用于可视化其图像表示分布。
首先，我们使用dentvfm提取每张图像的image-level表示。
然后我们使用tsne将高维表示降维到2维空间中，并用2维坐标系中的点表示。
属于不同图像类别的点被表示为不同的颜色。
为了进一步分析image-level的表示与图像不同patch的关联，我们可视化了最后一层tflayer中cls token的四个注意力头对应的self-attention map（MHSA）。
另外，我们还可视化了所有注意力头对应的self-attention map的融合，称为merged MHSA。
来源于不同head的MHSA被赋予不同颜色。
我们对不同的放射图像类型执行了MHSA可视化。
同时，我们还对从不同训练阶段的模型获得的MHSAmap进行可视化以分析模型预训练过程中知识的积累过程。
\end{comment}
We adopt the [CLS] token output from the final transformer layer of DentVFM as the image-level representation.
First, DentVFM is employed to extract image-level features from each image in target datasets. 
These high-dimensional representations are then projected into a two-dimensional space using t-SNE\cite{maaten2008visualizing}, where each data point is plotted in a 2D coordinate system. 
Points corresponding to different image categories are color-coded to facilitate visual discrimination. 
To further investigate the contribution of local anatomical regions to image-level representations, we visualize the MHSA maps associated with the [CLS] token from the final transformer layer. 
Specifically, we display the attention maps of four individual attention heads, as well as their merged result, known as the merged MHSA map. 
Each attention head is assigned a distinct color to highlight its unique focus and contribution.
MHSA map visualizations are performed across various types of dental radiographic images. 
Additionally, to analyze the progression of knowledge acquisition during pre-training, we visualize MHSA maps extracted from models at different training stages, thereby revealing how attention patterns evolve over time.

\subsubsection*{Visualization of voxel-level representations}
\begin{comment}
我们选择了来源于分割数据集tf3的volume用于可视化，数据集中的每个volume都包含了颌面解剖结构的分割标注。
对于一个volume，我们首先使用dentvfm-3d提取每个patch的表示。
通过对path表示进行插值，我们可以得到每个voxel的表示。
我们应用tsne将voxel-level的高维表示降维到2维空间中，并可视化为2维坐标系中的点。
属于相同解剖结构的体素对应的点被赋予相同的颜色。
\end{comment}
To verify awareness of local anatomical structures, we utilize volumetric data from segmentation datasets (e.g., \textit{Oral Struct Seg (TF3)}), where each volume is annotated with masks of various oral and maxillofacial anatomical structures.
For each volume, patch-level representations are first extracted using DentVFM-3D. 
These patch embeddings are then interpolated to generate voxel-level representations across the entire volume.
The high-dimensional voxel-level features are projected into a two-dimensional space using t-SNE. 
The resulting 2D embeddings are plotted as points in a Cartesian coordinate system, with voxels belonging to the same anatomical structure color-coded identically.
This allows for intuitive comparison of feature distributions across different anatomical regions.

\subsubsection*{Visualization of pixel-level representations}
\begin{comment}
我们使用dentvfm-2d从2d牙科放射图像中提取patch表示，使用dentvfm-3d从3d牙科放射图像中提取patch表示。
随后，patch级别的表示被插值到和输入图像同样的分辨率以获得pixel-level的表示。
对于volume，我们取一些2d slices进行可视化并不以3d的形式可视化。
我们使用k-means算法对pixel-level表示进行无监督的聚类。
被聚类为同一类别的pixel被赋予相同的颜色。
我们可视化和原图同样分辨率的k-means图。
\end{comment}
We use DentVFM-2D and DentVFM-3D to extract patch-level representations from 2D and 3D dental radiographic images, respectively. 
These patch-level representations are subsequently interpolated to the original image resolution, producing pixel-level (for 2D images) or voxel-level (for 3D images) representations. 
For 3D volumes, instead of visualizing the entire volume, we select representative slices for analysis and display. 
Unsupervised clustering is then performed on pixel-level representations using the k-means algorithm. 
The pixels assigned to the same cluster are visualized in identical colors, allowing intuitive identification of semantically similar regions. 
The clustering results are rendered as 2D maps aligned with the resolution of the original images, enabling direct visual comparison and interpretation of local structural patterns.

\subsection*{Evaluation and Statistical Analysis}

In our evaluation, we use the accuracy (ACC) as metric for classification tasks, and Dice and IoU as metrics for segmentation tasks. 
When segmentation tasks involve multiple semantic categories, we compute the mean Dice (mDice) and mean IoU (mIoU). 
We use Mean Radial Error (MRE) and Success Detection Rate (SDR) to evaluate dental landmark detection task.
For each task, we perform 5 random train-test splits and calculate the mean and standard deviation across the 5 iterations. 
We perform a two-tailed t-test to compare DentVFM with the most competitive task-specific models and clinical evaluations to determine if there are significant differences.

\subsection*{Computing Hardware and Software}
We use Python (v3.10.12) and Pytorch\cite{goldberger2000physiobank} (v2.4.0) for pre-training and evaluation. 
We reference the original DINOv2 algorithm (\url{https://github.com/facebookresearch/dinov2}) to implement our pre-training algorithm and evaluation framework. 
Model pretraining is conducted on the $2\times 8$ H100-SXM GPU nodes, utilizing Fully Sharded Data Parallel (FSDP) for distributed multi-GPU training. 
All downstream task fine-tuning is performed on a single H100-SXM GPU. 
We implement the logistic regression module using cuML (\url{https://github.com/rapidsai/cuml}). 
For 2D image segmentation, we implement the Mask2Former head and ViTAdapter using MMSegmentation (\url{https://github.com/open-mmlab/mmsegmentation}). 
For 3D image segmentation, we implement the UNETR head, following the original UNETR implementation (\url{https://github.com/tamasino52/UNETR}). 
In addition, we develop the 3D adapter manually. 
To ensure a fair comparison, we integrate the 3D segmentation model within the nnUNet framework (\url{https://github.com/MIC-DKFZ/nnUNet}) for the evaluation of 3D image segmentation.  
The pre-trained model weights used for comparison are obtained from the official pre-trained checkpoints, which can be found at the following link: Resnet50\cite{he2016deep} (\url{https://github.com/qubvel/segmentation_models}), SAM\cite{kirillov2023segment} (\url{https://github.com/facebookresearch/segment-anything}), SAM\_Med2d\cite{cheng2023sammed2d} (\url{https://github.com/OpenGVLab/SAM-Med2D}), SAM\_Med3d\cite{wang2024sam} (\url{https://github.com/uni-medical/SAM-Med3D}), SwimUNETR\cite{tang2022self} (\url{https://github.com/Project-MONAI/research-contributions/tree/main/SwinUNETR/Pretrain}),
CLIP\cite{radford2021learning} (\url{https://huggingface.co/openai/clip-vit-base-patch16}), BiomedCLIP\cite{zhang2023biomedclip} (\url{https://huggingface.co/microsoft/BiomedCLIP-PubMedBERT_256-vit_base_patch16_224}), M3D\cite{bai2024m3d} (\url{https://huggingface.co/GoodBaiBai88/M3D-CLIP}),
DINOv2\cite{oquab2023dinov2} (\url{https://github.com/facebookresearch/dinov2}), LVM-Resnet50\cite{mh2023lvm} (\url{https://github.com/duyhominhnguyen/LVM-Med}), LVM-ViT\cite{mh2023lvm} (\url{https://github.com/duyhominhnguyen/LVM-Med}).
The implementation of the task-specific models for comparison is available at the following link: MLUA\cite{wang2023multi} (\url{https://github.com/Zzz512/MLUA}).
LCD-Net\cite{hu2021location} is implemented by ourselves.

\subsection*{Ethics Statement}

This study was approved by the Ethics Committee of Shanghai Ninth People's Hospital. 
Only de-identified retrospective data are used for research, without the active involvement of patients.
% The requirement for ethical approval was waived by the Ethics Committee of Shanghai Ninth People's Hospital because this was a retrospective study and all data were fully anonymized. 

\section*{Acknowledgements}

We sincerely thank all medical professionals and data providers (Shanghai Ninth People’s Hospita, FUSSEN and other collaborators) for establishing the DentVista dataset and the DentBench evaluation framework. 
We acknowledge with gratitude the computational support provided by our collaborative institution (China Mobile Shanghai). 
This work was supported by the Shanghai Municipal Special Fund for Promoting High-Quality Industrial Development (2024-GZL-RGZN-02033)

\section*{Author Contributions Statement}

Study conceptualization: X.H., X.F., D.H., X.W., X.Z., S.Z.; 
Data acquisition, analysis, and interpretation: X.H., X.F., D.H., A.G.; 
Methodological design and implementation: X.H., X.F.; 
Statistical analyses: X.H., X.F., H.D.; 
Code and reproducibility: X.H., X.F., D.L.; 
Technical support: X.Z.; 
Writing of the manuscript: X.H.; 
Critical revision of the manuscript: All authors; 
Study supervision: X.W., X.Z., S.Z.

\section*{Data and Code Availability}

The DentVista dataset comprises clinical data sourced from multiple collaborative institutions. 
Due to its sensitive nature and contractual obligations with our partners regarding data access protocols, DentVista will remain confidential. 
For more information on DentVista, please contact X.H. 
DentBench is publicly accessible to facilitate the future development of AI models in dentistry. 
DOIs and links for the external data used in DentBench can be found in the Supplementary Table \ref{tab:dataset-summary}. 
The internal data used in DentBench can be accessed through a structured application process. 
% Access procedures are available at \hl{[link]}. 
The code used to train, fine-tune and evaluate DentVFM will be public to encourage continued advancement in the dental foundation model and broader community engagement and collaborative innovation.  
The model weights will be made available upon acceptance on paper. 

\clearpage

% \begin{figure}[ht]

% \begin{figure}[ht]

% \begin{figure}[ht]

% \begin{figure}[ht]

% \begin{figure}[ht]

% \begin{figure}[ht]

% reference
\bibliography{sample}

\begin{landscape}
\section*{Supplementary Materials}

% 定义 X 列（自动分配宽度）
\newcolumntype{Z}{>{\centering\arraybackslash}p{0.25\linewidth}}
\newcolumntype{S}{>{\centering\arraybackslash}p{0.15\linewidth}}
% 全局行距 & 列距调小
\renewcommand\arraystretch{1.05}
\setlength{\tabcolsep}{2.5pt} % 默认 6pt，这里缩小
\captionsetup[table]{name=Supplementary Table}
% \begin{landscape}
{\small % 可以改为 \footnotesize / \scriptsize 进一步压缩
\begin{tabularx}{\linewidth}{Y Y Z Y Y Y Y S}
\caption{Comprehensive overview of downstream tasks in DentBench, presenting their names, abbreviations, definitions, types, subspecialties, modalities, data sizes, and sources. In the source, "Public" denotes data obtained from publicly available datasets, whereas "Complementary" refers to additional datasets that we have curated.}\label{tab:dataset-summary} \\
\toprule
\makecell{\textbf{Task Name}} & 
\textbf{Abbreviation} & 
\textbf{Task Definitions} & 
\makecell{\textbf{Task Type}} & 
\textbf{Subspecialty} & 
\textbf{Modality} & 
\makecell{\textbf{Data Size}} & 
\textbf{Source} \\
\midrule
\endfirsthead
\toprule
\makecell{\textbf{Task Name}} &
\textbf{Abbreviation} & 
\textbf{Task Definitions} & 
\makecell{\textbf{Task Type}} &
\textbf{Subspecialty} &
\textbf{Modality} &
\makecell{\textbf{Data Size}} &
\textbf{Source} \\
\midrule
\endhead
\midrule
\multicolumn{7}{r}{\small\itshape Continued on next page} \\
\midrule
\endfoot
\bottomrule
\endlastfoot
Oral Abnormality Recognition &OAR (DENTEX) &Classifying cropped panoramic image regions into 4 dental abnormality types: caries, deep caries, impacted tooth, and periapical lesion &Dental Disease Diagnosis &Preventive Dentistry  &Panoramic X-ray  &632 &Public (\href{https://doi.org/10.5281/zenodo.7812323}{DENTEX}\cite{hamamci2023dentex}) \\

Oral Abnormality Recognition &OAR (DXPD) &Classifying cropped panoramic image regions into 22 dental abnormality types: caries, implant, missing teeth, bone loss, cyst etc. &Dental Disease Diagnosis & Preventive Dentistry &Panoramic X-ray  &1733 &Public (\href{https://www.kaggle.com/datasets/lokisilvres/dental-disease-panoramic-detection-dataset}{DXPD}) \\

Oral Abnormality Recognition &OAR (DRAD) &Classifying cropped panoramic image regions into 4 dental abnormality types: caries, fillings, impacted tooth, and implant &Dental Disease Diagnosis & Preventive Dentistry & Panoramic X-ray  &1992 &Public (\href{https://www.kaggle.com/datasets/imtkaggleteam/dental-radiography}{DRAD})\\

% Fracture简称Fx Diagnosis简称Dx
Cranio-maxillofacial Fracture Diagnosis &CMFFxDx (PAN) &Classification of CMF fracture locations in panoramic radiographs, including condyle, maxilla, mandible, and multiple facial sites &Dental Disease Diagnosis &Oral and Maxillofacial Surgery &Panoramic X-ray &544 &Complementary (NineH-CMFFx-PAN)\\

Cranio-maxillofacial Fracture Diagnosis &CMFFxDx (CT/CBCT) &Classification of CMF fracture locations based on CT or CBCT scans, including condyle, maxilla, mandible, and multiple facial sites &Dental Disease Diagnosis &Oral and Maxillofacial Surgery  &CT/CBCT  &286 &Complementary (NineH-CMFFx-CT/CBCT)\\

Cyst Diagnosis &CystDx &Classification of cystic lesions in panoramic radiographs into 4 categories: ameloblastoma, dentigerous cyst, keratocyst, and periapical cyst &Dental Disease Diagnosis & Oral and Maxillofacial Surgery &Panoramic X-ray   & 606 &Complementary (NineH-CystDx)\\

TMJ Abnormality Diagnosis &TMJADx (MRI) &Diagnosing disc displacement and changes in condylar position from TMJ MRI images, classifying them into 2 categories: normal and abnormal &Dental Disease Diagnosis &Oral and Maxillofacial Surgery  &MRI &240 &Complementary (NineH-TMJADx-MRI)\\

TMJ Abnormality Diagnosis &TMJADx (PAN) &Diagnosing disc displacement and changes in condylar position from panoramic X-rays, classifying them into 2 categories: normal and abnormal &Dental Disease Diagnosis &Oral and Maxillofacial Surgery &Panoramic X-ray  &401 &Complementary (NineH-TMJADx-PAN)\\

Malocclusion Diagnosis &MALODx &Classifying malocclusion types from lateral cephalometric radiographs, including skeletal Class I, Class II, and Class III &Dental Disease Diagnosis &Orthodontics  &Lateral X-ray  &336 &Public (\href{https://www.codabench.org/competitions/2576/}{CL-Detection}\cite{wang2016benchmark})\\

Coarse-grained Periodontal Grading &CGPerioG &Classification of periodontitis severity based on panoramic radiographs, categorized into 4 grades (1 to 4) &Dental Disease Diagnosis &Periodontics &Panoramic X-ray  &862 &Complementary (NineH-CGPerioG)\\

Fine-grained Periodontal Grading & FGPerioG &Classification of periodontitis severity based on cropped regions of panoramic radiographs, divided into 4 grades (1 to 4) &Dental Disease Diagnosis &Periodontics  &Panoramic X-ray   &2000 &Complementary (NineH-FGPerioG)\\

Caries Assessment & CarA & Classifying the severity of dental caries from panoramic radiographs into 3 categories: mild, moderate, and severe & Dental Disease Diagnosis &Endodontics  & Panoramic X-ray  & 2400 &Public (\href{https://github.com/Zzz512/MLUA}{DC1000}\cite{wang2023multi})\\
 
% Landmark Detection& Landmark Det.& Treatment Planning and Prognosis &Orthognathic Surgery  &Lateral Xray   & 100&Public
% (CL-Detection2024)\\

Orthognathic Surgery Treatment Planning & OSTP (LAT) & Classifying the type of orthognathic surgery from lateral cephalometric radiographs into single-jaw and double-jaw surgery & Dental Treatment Analysis
 & Orthognathic Surgery & Lateral X-ray  &297 &Complementary (NineH-OSTP-LAT)\\

Orthognathic Surgery Treatment Planning &OSTP (CT/CBCT) &Classifying the type of orthognathic surgery from CB or CBCT into single-jaw and double-jaw surgery &Dental Treatment Analysis  &Orthognathic Surgery  &CT/CBCT  & 156 &Complementary (NineH-OSTP-CT/CBCT)\\

Orthognathic Surgery Postoperative Analysis & OSPA (LAT) &Identifying the type of orthognathic surgery performed from postoperative lateral cephalometric radiographs, including single-jaw and double-jaw surgery &Dental Treatment Analysis & Orthognathic Surgery &Lateral X-ray  & 1077 &Complementary (NineH-OSPA-LAT)\\
 
Orthognathic Surgery Postoperative Analysis & OSPA (CT/CBCT) &Identifying the type of orthognathic surgery performed from postoperative CT or CBCT, including single-jaw and double-jaw surgery &Dental Treatment Analysis  &Orthognathic Surgery  &CT/CBCT  & 201 &Complementary (NineH-OSPA-CT/CBCT)\\

Segmental Orthognathic Treatment Planning & SOTP (CT/CBCT) &Classifying whether orthognathic segmentation is required based on CT or CBCT scans. &Dental Treatment Analysis & Orthognathic Surgery &CT/CBCT  &200 &Complementary (NineH-SOTP-CT/CBCT)\\

Segmental Orthognathic Postoperative Analysis & SOPA (CT/CBCT) &Classification whether orthognathic segmentation surgery was performed based on postoperative CT or CBCT scans &Dental Treatment Analysis  & Orthognathic Surgery &CT/CBCT  &206 &Complementary (NineH-SOPA-CT/CBCT)\\

% Implant PBR Prediction& PBR Pred.& Treatment Planning and Prognosis &Oral Implantology  &Panoramic Xray   & 100&In-house (NineH-RBP)\\

Bone Mineral Density Grading &BMDG &Grading bone density from panoramic radiographs into four levels based on the Lekholm and Zarb (L\&Z)\cite{lee2024deep} classification &Biomarker Identification & Oral Implantology &Panoramic X-ray   &1375 &Complementary (NineH-BMDG)\\

Development Assessment & DevA (PAN) &Estimation of physiological age based on panoramic radiographs in patients aged 6 to 20 years &Biomarker Identification & Pediatric Dentistry & Panoramic X-ray  & 1486 &Complementary (NineH-DevA-Pan)\\

Development Assessment & DevA (LAT) &Estimation of physiological age based on lateral cephalometric radiographs in patients aged 6 to 20 years &Biomarker Identification & Pediatric Dentistry & Lateral X-ray  & 1730 &Complementary (NineH-DevA-Lat)\\

Cephalometry Landmark Detection & CL-Detect &Accurately locating 53 landmark points in the lateral X-ray image &Landmark Detection & Orthognathic Surgery & Lateral X-ray  & 446 &Public (\href{https://www.codabench.org/competitions/2576/}{CL-Detection}\cite{zhang2024deep})\\

Adult Tooth Segmentation &ATS &Binary segmentation of adult teeth from panoramic radiographs &Lesion\&Anatomy Segmentation & Oral and Maxillofacial Radiology &Panoramic X-ray   &2000 &Public (\href{https://echoyq.github.io/ToothFairlySSS}{STS2D}\cite{bolelli2023tooth})\\

Children Tooth Segmentation &CTS  &Binary segmentation of children teeth from panoramic radiographs &Lesion\&Anatomy Segmentation &Oral and Maxillofacial Radiology  & Panoramic X-ray  &193 &Public (\href{https://doi.org/10.6084/m9.figshare.c.6317013.v1}{CDPRD}\cite{zhang2023children})\\

Tooth Segmentation and Labeling &TS\&L &Segmenting individual teeth from panoramic radiographs and labeling them with corresponding FDI numbers &Lesion\&Anatomy Segmentation  &Oral and Maxillofacial Radiology  &Panoramic X-ray   &2066 &Public (\href{https://www.kaggle.com/datasets/zwbzwb12341234/a-dual-labeled-dataset}{ADLD})\\

Fine-grained Tooth Segmentation &FGTS  &Fine-grained segmentation of each tooth, dentin, pulp, dental materials, and decay from cropped panoramic radiographs &Lesion\&Anatomy Segmentation  & Oral and Maxillofacial Radiology &Panoramic X-ray   & 26215 &Public (\href{https://github.com/Zzz512/TSD}{TSD-FG}\cite{wang2024dsis})\\

Coarse-grained Tooth Segmentation & CGTS &Coarse-grained segmentation of tooth, dentin, pulp, dental materials, and decay from whole panoramic radiographs &Lesion\&Anatomy Segmentation   & Oral and Maxillofacial Radiology & Panoramic X-ray  & 895 &Public (\href{https://github.com/Zzz512/TSD}{TSD-FG}\cite{wang2024dsis})\\
 
Mandible Segmentation &MS  &Segmentation of the maxilla and mandible from panoramic radiographs &Lesion\&Anatomy Segmentation  & Oral and Maxillofacial Radiology &Panoramic X-ray   & 116 &Public (\href{https://data.mendeley.com/datasets/hxt48yk462/2}{PXWSM}\cite{abdi2020panoramic})\\

Bitewing Segmentation &BS (DBXD)  &Semantic segmentation of bitewing radiographs into 15 classes, including bone, caries, crowns, implants, implant crowns, dentin, enamel, and others &Lesion\&Anatomy Segmentation   &Oral and Maxillofacial Radiology  & Bitewing X-ray &1099 &Public (\href{https://www.kaggle.com/datasets/shubhamskg/bitewing-datasets}{DBXD})\\

Bitewing Segmentation &BS (BK) &Semantic segmentation of abnormalities in bitewing radiographs, including crowns, implants, restorations, and root canal treatments &Lesion\&Anatomy Segmentation   &Oral and Maxillofacial Radiology  &Bitewing X-ray  & 271 &Public (\href{https://www.kaggle.com/datasets/shubhamskg/bitewing-datasets}{BK})\\

Anatomy Structure Segmentation & ASS (TF2) &Semantic segmentation of 42 anatomical structures in CBCT scans &Lesion\&Anatomy Segmentation  &Oral and Maxillofacial Radiology  & CBCT &480 &Public (\href{https://ditto.ing.unimore.it/toothfairy2/}{ToothFairy2}\cite{bolelli2025segmenting,bolelli2024segmenting,lumetti2024enhancing})\\

Anatomy Structure Segmentation & ASS (TF3) &Semantic segmentation of 77 anatomical structures in CBCT scans &Lesion\&Anatomy Segmentation   &Oral and Maxillofacial Radiology  &CBCT  & 532 &Public (\href{https://ditto.ing.unimore.it/toothfairy3/}{ToothFairy3}\cite{bolelli2025segmenting,bolelli2024segmenting,lumetti2024enhancing})\\
 
Tooth Segmentation & TS (sts3d) &Binary segmentation of adult teeth from CBCT &Lesion\&Anatomy Segmentation  & Oral and Maxillofacial Radiology & CBCT &30 &Public (\href{https://echoyq.github.io/ToothFairlySSS/}{STS3D})\\
 
Tooth Segmentation &TS (NC)  &Binary segmentation of teeth in CBCT scans &Lesion\&Anatomy Segmentation   &Oral and Maxillofacial Radiology  &CBCT  & 148 &Public (NC\cite{cui2022fully})\\

Head\&Neck Structure Segmentation & H\&NS &Segmentation of left and right parotid glands, brainstem, left and right optic nerves, mandible, and left and right submandibular glands from CT images of radiotherapy tumor patients &Lesion\&Anatomy Segmentation & Oral and Maxillofacial Radiology &CT  &48 &Public (\href{https://www.imagenglab.com/newsite/pddca/}{Head\&Neck}\cite{raudaschl2017evaluation})\\

Pulpy Segmentation & PulpyS &Segmentation of 19 classes in CBCT scans, including the inferior alveolar canal, lower teeth, and abnormal teeth &Lesion\&Anatomy Segmentation & Oral and Maxillofacial Radiology &CT  & 443 &Public (\href{https://ditto.ing.unimore.it/pulpy3d/}{Pulpy3D}\cite{gamal2024automatic})\\

Caries Segmentation &CarS  &Segmentation of caries from cropped panoramic image regions &Lesion\&Anatomy Segmentation  &Oral and Maxillofacial Radiology  & Panoramic X-ray  &2400 &Public (\href{https://github.com/Zzz512/MLUA}{DC1000}\cite{wang2023multi})\\
 
Oral Abnormal Segmentation &OAS  &Segmentation of abnormal regions from panoramic radiographs &Lesion\&Anatomy Segmentation  &Oral and Maxillofacial Radiology  & Panoramic X-ray  & 119 &Public (\href{https://tdd.ece.tufts.edu/}{Tufts}\cite{panetta2021tufts})\\
\end{tabularx}
}
\end{landscape}

\captionsetup[table]{name=Supplementary Table}
\begin{table}[H]
\centering
\caption{A detailed description of the composition of DentVista. DentVista is composed of data from three sources, primarily originating from the East Asia region.}
\label{tab:dentvista_sources}
\begin{tabularx}{\textwidth}{X X l l}
\toprule
\textbf{Data Source} & \textbf{Modality} & \textbf{Number of Images} & \textbf{Region} \\
\midrule
3 Hospitals\newline(Private) & Panoramic X-ray, Intraoral X-ray, Lateral X-ray, Anteroposterior X-ray, MRI, CBCT, CT & 699,429 & Chinese Mainland \\
\addlinespace
105$\times$Dental Clinics\newline(Private) & Panoramic X-ray, Intraoral X-ray, Lateral X-ray, CBCT & 938,997 & Chinese Mainland \\
\addlinespace
Web Data\newline(\href{https://zenodo.org/records/4457648}{Zenodo}, \href{https://data.mendeley.com/datasets/73n3kz2k4k/2}{Mendeley}, \href{https://humansintheloop.org/resources/datasets/teeth-segmentation-dataset/}{Humansintheloop}) & Panoramic X-ray & 1,705 & Paraguay, Tunisia, Congo \\
\bottomrule
\end{tabularx}
\end{table}

\setcounter{table}{2}
\captionsetup[table]{name=Supplementary Table}
\begin{table}[H]
\centering
\caption{Architecture details of the ViT-B/L/G networks used in this work. We use a patch size of 14 for DentVFM-2D, while use a patch size of 16 for DentVFM-3D. We employ SwiGLU\cite{shazeer2020glu} as the FFN layer.}
\label{tab:model_architecture}
\begin{tabularx}{\textwidth}{l c c c c}
\toprule
\textbf{Architecture} & \textbf{Embed dim} & \textbf{Heads} & \textbf{Blocks} & \textbf{Params} \\
\midrule
ViT-B & 768 & 12 & 12 & $\approx86M$ \\
\addlinespace
ViT-L & 1024 & 16 & 24 & $\approx307M$ \\
\addlinespace
ViT-G & 1536 & 24 & 40 & $\approx1.1B$ \\
\bottomrule
\end{tabularx}
\end{table}

\setcounter{table}{3}
\captionsetup[table]{name=Supplementary Table}
\begin{table}[H]
\centering
\caption{A detailed description of baselines. We carefully select the methods for comparison based on differences in data domains and learning paradigms.}
\label{tab:baseline_details}
\begin{tabularx}{\textwidth}{l c c c c}
\toprule
\textbf{Dim} & \textbf{Data Domain} & \textbf{Learning Paradigm} & \textbf{Baseline} & \textbf{Data Size} \\
\midrule
\multirow{7}{*}{2D} 
    & \multirow{4}{*}{Natural Domain} 
       & \multirow{2}{*}{Supervised Learning} 
          & Resnet50\cite{he2016deep} & 1.28M \\
    &  &  & SAM\cite{kirillov2023segment} & 1B \\
    &  & Weakly Supervised Learning & CLIP\cite{radford2021learning} & 400M \\
    &  & Self-supervised Learning & DINOv2\cite{oquab2023dinov2} & 142M \\
\addlinespace
    & \multirow{3}{*}{Medical Domain} 
        & Supervised Learning & SAM\_Med2d\cite{cheng2023sammed2d} & 4.6M \\
    &  & Weakly Supervised Learning & BiomedCLIP\cite{zhang2023biomedclip} & 15M \\
    &  & Self-supervised Learning & LVM-ViT\&Resnet50\cite{mh2023lvm} & 1.3M \\
\midrule
\multirow{3}{*}{3D} 
    & \multirow{2}{*}{Medical Domain} 
        & Supervised Learning & SAM\_Med3d\cite{wang2024sam} & 140K \\
    &  & Supervised Learning & SwimUNETR\cite{tang2022self} & 5K \\
    &  & Weakly Supervised Learning & M3D\cite{bai2024m3d} & 120K \\
\bottomrule
\end{tabularx}
\end{table}

\setcounter{table}{4}
\captionsetup[table]{name=Supplementary Table}
\begin{table}[H]
\centering
\caption{Detailed hyper-parameter configurations for pre-training.}
\label{tab:hyperparameters}
\begin{tabularx}{\textwidth}{l *{3}{>{\centering\arraybackslash}X}}
\toprule
\textbf{Hyper-parameter} & \textbf{DentVFM-2DB} & \textbf{DentVFM-2DL/G} & \textbf{DentVFM-3DB/L/G} \\
\midrule
Stochastic drop path rate & 0.3 & 0.4 & 0.3 \\
Global crop size \& number & 224 \& 2 & 224 \& 2 & 96 \& 2 \\
Local crop size \& number & 98 \& 8 & 98 \& 8 & 48 \& 8 \\
Dino head prototypes \& dim & 65536 \& 256 & 131072 \& 384 & 65536 \& 256 \\
iBoT head prototypes \& dim & 65536 \& 256 & 131072 \& 256 & 65536 \& 256 \\
Masking ratio & (0.1, 0.5) & (0.1, 0.5) & (0.1, 0.5) \\
Shared head & False & False & False \\
Batch size & 2048 & 1024 & 1024 \\
Total iterations & 125000 & 625000 & 90000 \\
Warmup iterations & 12500 & 100000 & 3000 \\
Learning rate (start-peak-final) & (0, 0.001, 1.0e-06) & (0, 0.0002, 1.0e-06) & (0, 0.0002, 1.0e-06) \\
Weight decay (start-final) & 0.04 & 0.04 & 0.04 \\
\bottomrule
\end{tabularx}
\end{table}

\end{document}